\documentclass[conference,letterpaper]{IEEEtran}
\IEEEoverridecommandlockouts

\usepackage[utf8]{inputenc} 
\usepackage[T1]{fontenc}    
\usepackage{url}            
\usepackage{booktabs}       
\usepackage{amsfonts}       
\usepackage{nicefrac}       
\usepackage{microtype}      
\usepackage{xcolor}         
\usepackage[cmex10]{amsmath}

\usepackage{graphicx} 
\usepackage{caption}
\usepackage{subfigure}
\usepackage{makecell}
\usepackage{framed}  
\usepackage{amsmath,amsfonts,amssymb,amsthm,epsfig,epstopdf,url,array}
\usepackage{bm}     
\usepackage{cite}
\usepackage{algorithm} 
\usepackage{algorithmic}
\usepackage{color}
\usepackage{xparse}
\usepackage{mathtools}     
\usepackage{wasysym}
\usepackage[flushleft]{threeparttable}
\newtheorem{theorem}{Theorem}

\newtheorem{proposition}{Proposition}
\newtheorem{assumption}{Assumption}

\def\BibTeX{{\rm B\kern-.05em{\sc i\kern-.025em b}\kern-.08em
    T\kern-.1667em\lower.7ex\hbox{E}\kern-.125emX}}

\usepackage{amsmath,amsfonts,bm}

\def\eqref#1{equation~\ref{#1}}

\def\1{\bm{1}}

\DeclareMathAlphabet{\mathsfit}{\encodingdefault}{\sfdefault}{m}{sl}
\SetMathAlphabet{\mathsfit}{bold}{\encodingdefault}{\sfdefault}{bx}{n}

\begin{document}


    \makeatletter
    \newcommand{\linebreakand}{%
      \end{@IEEEauthorhalign}
      \hfill\mbox{}\par
      \mbox{}\hfill\begin{@IEEEauthorhalign}
    }
    \makeatother

\title{SplitGP: Achieving   Both    Generalization and Personalization in Federated Learning }
   \author{
   \IEEEauthorblockN{Dong-Jun Han} 
    \IEEEauthorblockA{ 
      Purdue Univirsity \\ 
    han762@purdue.edu
    }
    \and
   \IEEEauthorblockN{Do-Yeon Kim}
    \IEEEauthorblockA{ 
     KAIST \\
    dy.kim@kaist.ac.kr}
    \and 
   \IEEEauthorblockN{Minseok Choi}
    \IEEEauthorblockA{ 
    Kyung Hee University \\
    choims@khu.ac.kr}
\and
   \IEEEauthorblockN{Christopher G. Brinton}
    \IEEEauthorblockA{ 
    Purdue University \\
    cgb@purdue.edu}
    \and
   \IEEEauthorblockN{Jaekyun Moon}
    \IEEEauthorblockA{ 
      KAIST \\
   jmoon@kaist.edu}
   
   \thanks{To appear in IEEE INFOCOM 2023.}
 }
\maketitle
\pagestyle{plain}

\begin{abstract}
 A fundamental challenge to providing edge-AI services is the need for a machine learning (ML) model that achieves personalization (i.e., to individual clients) and generalization (i.e., to unseen data) properties concurrently. Existing techniques in federated learning (FL) have encountered a steep tradeoff between these objectives and impose large computational requirements on edge devices during training and inference. In this paper, we propose SplitGP, a new split learning solution that can simultaneously capture generalization and personalization capabilities for efficient  inference across resource-constrained clients (e.g., mobile/IoT devices). Our key
idea is to split the full ML model into client-side and server-side 
components, and impose different roles to them:  the client-side model is trained to have strong personalization capability optimized to each client’s main task, while the server-side model is trained to have strong generalization capability for handling  all clients’ out-of-distribution tasks. We analytically characterize the convergence behavior of SplitGP, revealing that all client models approach stationary points asymptotically. Further, we analyze the  inference time in SplitGP and provide bounds for determining model split ratios. Experimental results show that SplitGP outperforms  existing baselines by wide margins in inference time and test accuracy for varying amounts of out-of-distribution samples.


\end{abstract} 

\section{Introduction}

With the increasing prevalence of mobile and Internet-of Things (IoT) devices, there is an explosion in demand for machine learning (ML) functionality across the intelligent network edge. From the service provider's perspective, providing a high-quality edge-AI service  to individual clients
is of paramount importance: given newly collected data, the goal of each client is to apply the provided ML model for inference/decisioning.  
However, there are two critical challenges that need to be handled to satisfy the client needs in practical edge-AI settings. 

\textbf{Issue 1: Personalization vs. generalization.} First, during the inference stage (i.e., after training has completed), each client should  be able to make reliable predictions not only  for  dominant data classes which have been observed locally,  but also occasionally  for the classes that have not previously appeared in its local data.  We refer to these as a client's main classes and out-of-distribution classes, respectively. Federated learning (FL) \cite{kairouz2021advances, li2020federated_survey, mcmahan2017communication}, the most recently popularized technique for distributing ML across edge devices, has demonstrated a sharp tradeoff between these objectives. In particular, existing works have aimed  to create either a   \textit{generalized global model} \cite{mcmahan2017communication, li2020federated,  wang2020federated,  acar2020federated,  karimireddy2020scaffold, wang2019adaptive, amiri2020federated, chen2020joint, park2021few,  park2021sageflow,   han2021fedmes,   yang2019scheduling, tu2020network, wang2021device, wang2020optimizing}
 that is tuned to the data distribution across all clients,   or \textit{personalized local models} \cite{smith2017federated, deng2020adaptive, fallah2020personalized, zhang2020personalized, li2021ditto} that work   case-by-case on each client's individual data.
  For example, a global activity recognition classifier for wearables learned via FL would be optimized for classes of activities observed over all users.     We term the capability to classify all classes as ``generalization''.   
  The generalized global model is a good option when the input data distribution appearing at each client during inference resembles the global training distribution.  
However, when the data distributions across clients are significantly non-IID (independent and identically distributed),   the globally aggregated FL model may not be the best option for many clients (e.g., consider activity sensors for individuals playing different types of sports). Personalized FL approaches tackle this
problem by providing a customized local model to each client  based on their individual local data distributions (e.g., a basketball vs. football player).  We term this capability to classify the local classes as ``personalization''. 

However, when a client needs to make   predictions for  classes that are not in its local data (i.e., due to distribution shift),   the personalized  FL model shows much lower performance than the generalized  model (see Sec. \ref{sec:exp}). Hence, it is important to capture both personalization (for handling local classes) and generalization (for handling out-of-distribution classes)    in practice where not only the main classes but also the out-of-distribution classes appear occasionally  during inference.   


\textbf{Issue 2: Inference requirements.} Mobile edge and IoT devices suffer from limited storage and computation resources. As a result, it is challenging to deploy   large-scale models (e.g., neural networks with millions of parameters)  at individual clients  for inference tasks without incurring significant costs.  Deploying the full  model at a nearby edge server can be another option, but this approach requires direct transmissions of raw data from the client during inference, and can also incur noticeable latency. Moreover, under this framework, when client models are personalized, the server would need to store all of these variations, which presents scalability challenges.

These two issues are thus significant obstacles to high quality edge-AI services, with existing approaches falling short of addressing them simultaneously. Motivated by this, we pose the following research question: \textit{How can we achieve both learning personalization and generalization across resource-constrained edge devices for high-quality inference?}




\begin{figure*}[t]
        \vspace{-3.5mm}
\centering
    \centerline{\includegraphics[width=166mm]{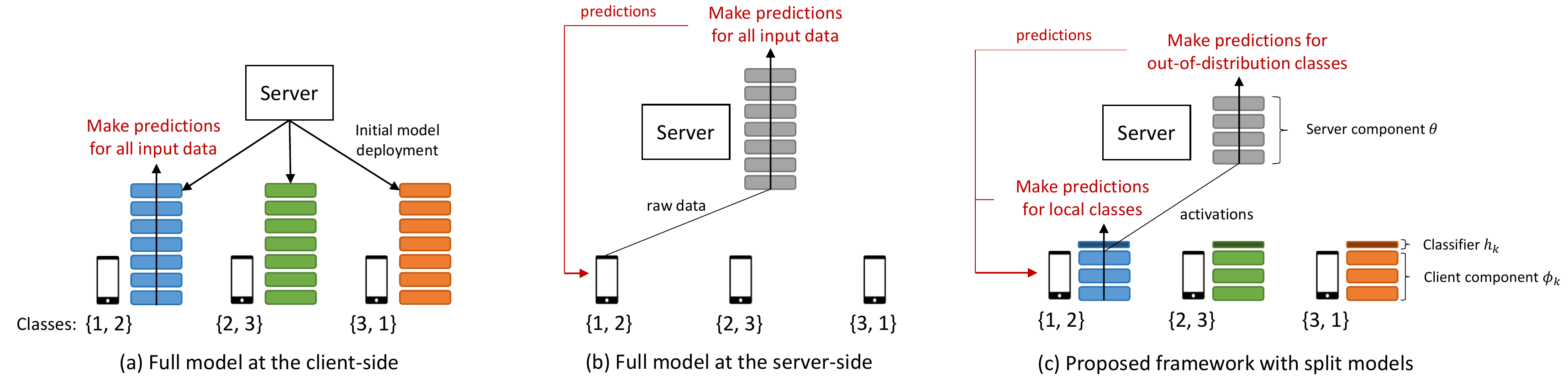}} 
        \vspace{-1.5mm}
    \caption{\small  \textbf{Comparison of inference  procedures:} Deploying the full model at individual clients (as in Fig. \ref{fig:inference_stage}(a)) is challenging in resource-constrained edge-AI settings as it induces significant storage and computational burden during inference.  When the model is implemented at the server (as in Fig. \ref{fig:inference_stage}(b)), raw data should be directly transmitted from the client to the server during inference, which incurs various privacy, communication and latency issues. 
The proposed framework  based on model splitting and edge computing  in Fig. \ref{fig:inference_stage}(c) captures both learning personalization and generalization requirements in resource-constrained scenarios while retaining desirable privacy and latency properties. }  
    \label{fig:inference_stage}
    \vspace{-4mm}
\end{figure*}

\textbf{Overview of approach.} 
To address this question, we  propose SplitGP, a new split learning (SL) solution for generalization and personalization in FL settings.           Our key idea is to split the full model  into  two parts, client-side and server-side, and  impose different roles to them during inference. The  client-side model should have strong \textit{personalization capability}, where the  goal is   to  work well on each user’s local distribution. On the other hand, the server-side model,  shared by all clients in the system, should have strong \textit{generalization capability} across the tasks of all users. During inference, each client solves its personalized task, i.e., for its main classes,  using the  client-side model. When the client has to make a prediction that is not related to its personalized task, i.e., for out-of-distribution classes, they can send the output feature from the client-side model to the server-side model, and receive the predicted result back from the edge server. 
We will show that this combination of model splitting and edge computing captures both personalization and generalization while  reducing the storage, computational load and latency during inference compared to existing methods where the full model is provided to individual clients. SplitGP also has significant advantages in terms of privacy and latency compared to schemes where the full model is deployed at the server-side.     Fig. \ref{fig:inference_stage} compares the inference stage of  SplitGP with these  existing frameworks; note that the models in Figs.  \ref{fig:inference_stage}(a)\&(b) are realized through existing FL and SL methodologies.  


\vspace{-0.2mm}

\textbf{Summary of contributions.} To the best of our  knowledge, simultaneous training of client/server-side models  with  different roles (personalization and generalization) has not been considered before. Existing works in distributed ML also tend to focus on the training process, without considering   the inference stage of clients with small storage space  and small computing powers. Overall, our   contributions are summarized as follows:
\vspace{-0.3mm}
\begin{itemize}
\item We propose SplitGP (Sec. \ref{sec:mainalgo}), a hybrid federated and split learning solution  that captures both learning generalization and personalization needs with multi-exit neural networks  for inference at resource-constrained clients.
\item We analytically characterize the convergence behavior of SplitGP (Sec. \ref{sec:convergence}), showing that training at each client will converge to a stationary point asymptotically under common assumptions in distributed ML.
\item We conduct a latency analysis of SplitGP (Sec. \ref{sec:latency}), which leads to guidelines on model splitting and insights on the power-rate regime
where our scheme is beneficial.
\end{itemize}
Experimental results (Sec. \ref{sec:exp}) show that SplitGP  outperforms existing approaches in practical scenarios with wide margins of improvement in testing accuracy and inference time. 

\section{ Related Works}\label{sec:rel}



 
\textbf{Federated learning.} A large number of FL techniques \cite{mcmahan2017communication, li2020federated,  wang2020federated,  acar2020federated,  karimireddy2020scaffold,wang2019adaptive, amiri2020federated, chen2020joint, yang2019scheduling, tu2020network, wang2021device, wang2020optimizing} including FedAvg \cite{mcmahan2017communication}, FedProx \cite{li2020federated}, FedMA \cite{wang2020federated}, FedDyn \cite{acar2020federated} and SCAFFOLD \cite{karimireddy2020scaffold} have been proposed  to construct  generalized global models. Personalized FL \cite{smith2017federated, deng2020adaptive, fallah2020personalized, zhang2020personalized, li2021ditto} has been studied more recently through techniques such as multi-task learning \cite{smith2017federated}, interpolation and finetuning \cite{deng2020adaptive}, meta-learning \cite{fallah2020personalized}, and regularization \cite{li2021ditto}. However, these strategies do not achieve generalization and personalization simultaneously, and thus are not the best options in practice  where not only the main classes and but also the out-of-distribution classes  appear occasionally during inference. 
In this respect, a recent work \cite{chen2021bridging} proposed simultaneously constructing generalized and personalized models in FL with two models sharing the same feature extractor. 
In practice,  the  full model would need to be deployed at each client for achieving personalization during inference, which is challenging in the edge-AI scenarios we consider with resource-constrained clients. 



\textbf{Split learning.}  
Recently, various SL schemes have been proposed   \cite{vepakomma2018split, gupta2018distributed, thapa2022splitfed, he2020group, han2021accelerating, oh2022locfedmix} to reduce  client-side storage and computation requirements during training compared to FL.  The  training process of our approach draws from concepts in SL  in that we divide the full model into client-side and server-side components.   However,  existing works  on   SL  including \cite{thapa2022splitfed, he2020group, han2021accelerating} do  not focus on capturing both generalization and personalization simultaneously.    Compared to  existing works, we consider a new inference scenario where the clients should make predictions frequently for the main classes but also occasionally for the out-of-distribution classes, and design a solution tailored to this setup.  We also let clients to frequently make predictions   using only the client-side model (instead of the full model), which results in reduced inference time.      


\textbf{Efficient edge-AI inference.}
Only a few prior works in distributed ML have focused on the inference stage of the clients at the edge.  \cite{teerapittayanon2017distributed} proposed to deploy distributed deep neural networks on the server and    devices during inference, using multi-exit neural networks \cite{teerapittayanon2016branchynet, hu2019learning, huang2018multi, li2019improved, phuong2019distillation}. Our approach also borrows the concept of  multi-exit neural networks with two exits  to make predictions   both  at the client-side and at the  server-side. However,  these previous works have assumed that the model training phase occurs in a centralized manner, which does not consider the important challenge of non-IID local datasets in FL/SL setups where raw data remains at the devices. In practical settings where each client observes main and out-of-distribution classes during inference, incorporating personalization and generalization results in significant performance enhancements during inference, as we will see in Sec. \ref{sec:exp}.

\vspace{-0.25mm}

\section{Proposed SplitGP Algorithm} \label{sec:mainalgo}
\vspace{-0.25mm}

 Let $K$ be the number of clients in the system and $D_k$ be the local dataset of  client $k = 1,2,\dots,K$ to be used for training an ML model. We denote the \textit{full model}  as a parameter vector $w$, which is split into \textit{client-side} and \textit{server-side} model components $\phi_k$ and $\theta$, respectively. Each client  also maintains an \textit{auxiliary classifier} $h_k$, with output dimension equal to the number of classes, which enables each client $k$ to make predictions using only $\phi_k$ and $h_k$; as shown in Fig. \ref{fig:inference_stage}(c), the output of $\phi_k$ becomes the input of $h_k$, and prediction can be made at the output of $h_k$. As in FL, model training will proceed in a series of training rounds, which we index $t = 0, 1...,T-1$.

Before training begins, we  split the initialized full model $w^0$ into  $w^0=[\phi^0,\theta^0]$, and also initialize $h^0$. Each client $k$ receives   $\phi^0, h^0$ and  sets $\phi_k^0 = \phi^0$, $h^0_k = h^0$, whereas $\theta^0$ is deployed at the server. 
After $T$ global rounds of training, each client $k$ obtains $\phi_k^T$, $h_k^T$, while the server obtains $\theta^T$.  We let
\vspace{-1mm}
\begin{equation}
v_k^t=[\phi_k^t, h_k^t, \theta^t]
\end{equation}
be the model components obtained at client $k$ and the server   when global round $t$ is finished.

\textbf{Inference scenario and goal.} We consider a   scenario having distribution shift between training and inference in each client: each client should make predictions mainly for the local classes but also occasionally for the out-of-distribution classes due to distribution shift. We  introduce a parameter for the relative portion of out-of-distribution  test samples, which is defined as
\begin{equation}\label{eq:rho}
\rho = \frac{\text{\# of out-of-distribution test samples}}{\text{\# of main test samples}}.
\end{equation}
Compared to the previous works on personalized FL focusing on  $\rho=0$, we consider a practical setup with $\rho>0$ caused by distribution shift between training and inference.  

As depicted in Fig. \ref{fig:inference_stage}(c),  the goal of the $k$-th client's model,  $\phi_k$ combined with $h_k$, is to  make a reliable prediction for local classes in $D_k$. The goal of each full model, $\phi_k$ combined with $\theta$, is to make make a reliable prediction for all classes in the network-wide dataset, $D=\cup_{k=1}^KD_k$, to handle the out-of-distribution classes of each client.  

\vspace{-0.25mm}

\subsection{Multi-Exit  Objective Function}
\vspace{-0.25mm}
Based on the three model  components $v=[\phi, h, \theta]$, we first define the following two losses computed based on $D_k$.

\textbf{Client-side loss.} Given $k$-th client's local data $D_k$ and $v=[\phi, h, \theta]$, the client-side loss $\ell_{C,k}(v)$ is defined as
\vspace{-1mm}
\begin{equation}\label{eq:client_loss}
 \ell_{C,k}(v)=  \frac{1}{|D_k|}\sum_{x\in D_k}\ell(x; \phi,h),
\end{equation}
where  $\ell(x;\phi,h)$ is the  loss (e.g., cross-entropy loss) computed with the  client  model ($\phi$ combined with $h$) using input data $x$.   (\ref{eq:client_loss}) is computed by client $k$.

\textbf{Server-side loss.} We also define the server-side loss $\ell_{S,k}(v)$ computed with the $k$-th client's local data $D_k$, as follows:
\vspace{-1mm}
\begin{equation}\label{eq:server_loss}
 \ell_{S,k}(v)= \frac{1}{|D_k|}\sum_{x\in D_k}\ell(x; \phi,\theta). 
\end{equation}
Here, $\ell(x; \phi,\theta)$ is the loss computed at the output of the full model ($\phi$ combined with $\theta$) 
 based on input  $x$.  As in existing SL schemes, (\ref{eq:server_loss}) is computed by the server in SplitGP. To facilitate this, for each $x \in D_k$, the client transmits the output features it computes from $\phi_k$ along with the label to the server\footnote{Potential privacy issues  can be handled by adding a noise layer \cite{lecuyer2019certified} at the client as in \cite{thapa2022splitfed}, which constructs private/noisy versions of output features.}.

\textbf{Proposed objective function.} In this way, the model $\phi$ maintained at the client-side affects both the client-side loss $\ell_{C,k}(v)$ and the server-side loss $\ell_{S,k}(v)$. By viewing the model $v=[\phi, h, \theta]$ as a multi-exit neural network \cite{teerapittayanon2016branchynet, hu2019learning, huang2018multi, li2019improved, phuong2019distillation} with two exits ($\ell_{C,k}$ and $\ell_{S,k}$), we update  $\phi$, $h$, $\theta$   to minimize the weighted sum of client/server-side losses computed with $D_k$:
\vspace{-1mm}
\begin{equation}\label{eq:sum_loss}
F_k(v) = \gamma \ell_{C,k}(v) + (1-\gamma) \ell_{S,k}(v).
\end{equation}
Here, $\gamma$ and $1-\gamma$ correspond to the weights of the client-side loss and the server-side loss, respectively.  If $\gamma=0$, the client-side model is updated only considering   the server-side loss, which corresponds to the objective function of  SplitFed proposed in \cite{thapa2022splitfed}. Personalization capability is not guaranteed at the client-side in this case.  
If $\gamma=1$, the client-side model does not consider the server-side loss at all, which does not guarantee generalization capability at the server-side. In multi-exit  network literatures \cite{huang2018multi, li2019improved, phuong2019distillation}, a common choice is to give equal weights\footnote{Our work can be combined with existing strategies that consider different weights for each exit's loss \cite{hu2019learning}   to further improve the  performance.} to both exits with $\gamma=0.5$.



\subsection{Personalization and Generalization  Training}

\textbf{Model update.} In the beginning of   global round $t$, we have $v_k^t=[\phi_k^t, h_k^t, \theta^t]$, where $\phi_k^t$ and $h_k^t$ are implemented at client $k$ while $\theta^t$ is deployed at the server.   Based on the proposed objective function  (\ref{eq:sum_loss}), the models of client $k$ $(\phi_k^{t}$ and $h_k^t$) and the shared server-side model $\theta^t$  are  updated according to 
\vspace{-1mm}
\begin{align}
\phi_k^{t+1}  =   \phi_k^{t}  -  \eta_t\tilde{\nabla}_{\phi} F_k(v_k^t),\label{eq:phi_update} \\
h_k^{t+1}  =   h_k^{t}  -  \eta_t\tilde{\nabla}_{h} F_k(v_k^t),\label{eq:h_update} \\
\theta_k^{t+1}  =   \theta^{t}  -  \eta_t\tilde{\nabla}_{\theta} F_k(v_k^t),\label{eq:theta_update}
\end{align}
where $\eta_t$  is the learning rate at global round $t$, and $\tilde{\nabla} F_k(v_k^t)=\frac{1}{|\tilde{D}_k^t|}\sum_{x\in \tilde{D}_k^t} \left(\gamma \nabla\ell_C(v;x) + (1-\gamma) \nabla\ell_S(v;x)\right)$ is the stochastic gradient computed with a specific mini-batch $\tilde{D}_k^t \subset D_k$. Fig. \ref{fig:update} shows the model update process at  client $k$. 


\textbf{Server-side model aggregation.} The updated server-side models based on (\ref{eq:theta_update}) are aggregated   
according to
$\theta^{t+1} \leftarrow  \sum_{i=1}^K\alpha_i\theta_i^{t+1}$,
to construct a single server-side model, where $\alpha_i = \frac{|D_i|}{\sum_{k=1}^K|D_k|}$ is the relative dataset size. This is a natural choice to capture  generalization capability at the server using a single model. 
\begin{figure}[t]
        \vspace{-2mm}
\centering
  \subfigure[Loss computation and parameter update process. The overall model can be viewed as a multi-exit neural network with two exits.]{\includegraphics[width=0.42\textwidth]{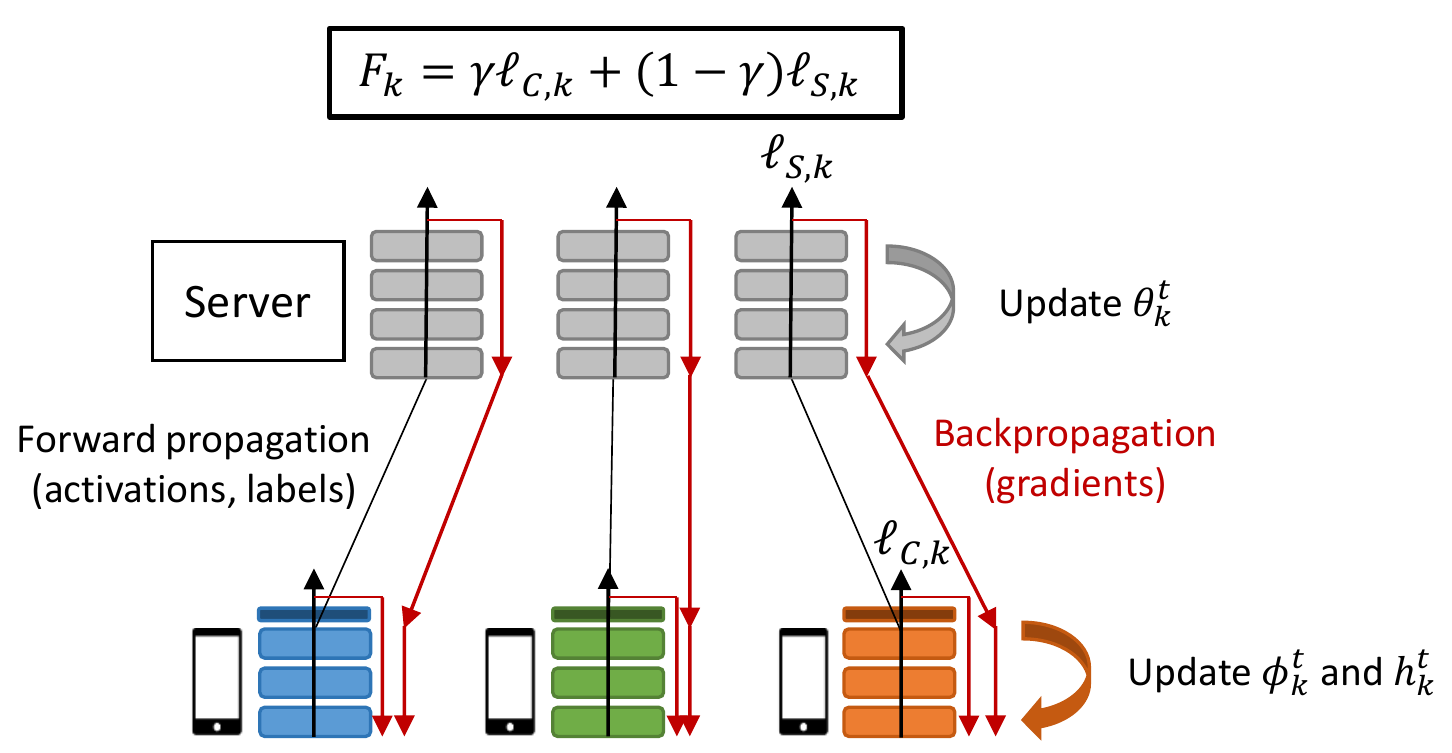}\label{fig:update}}
  \subfigure[Model aggregation process.]{\includegraphics[width =0.44\textwidth]{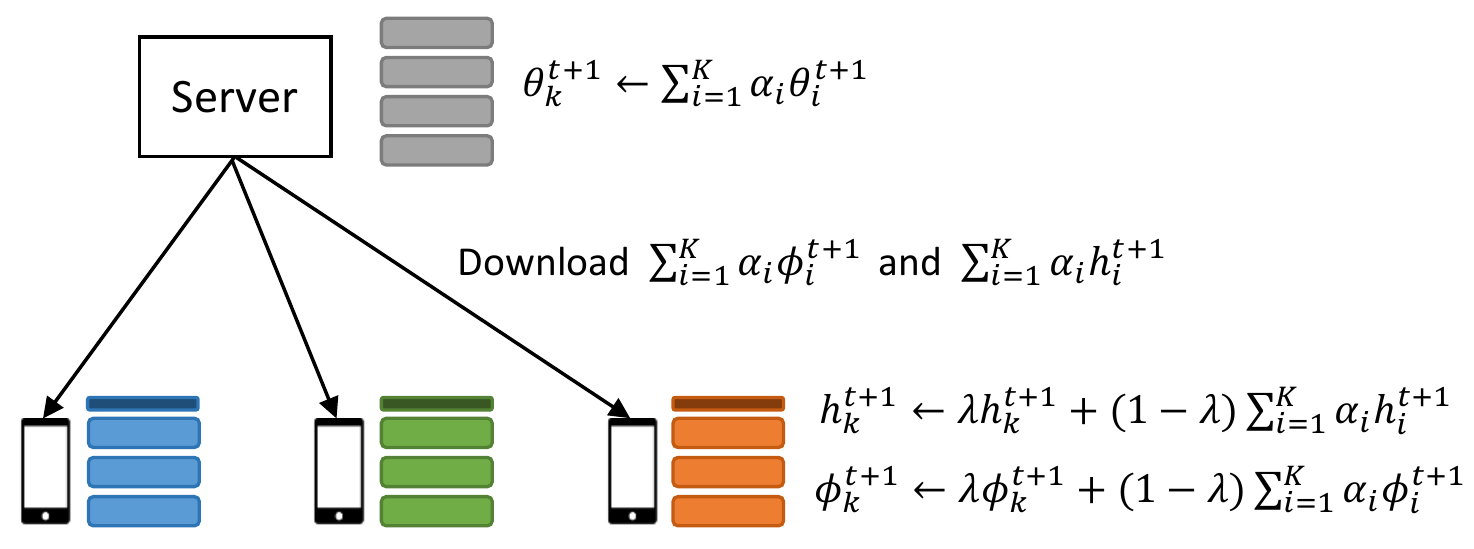}\label{fig:aggregation}}
        \vspace{-1mm}
  \caption{ \small Training process of   SplitGP at global round $t$. 
  After updating the models  using $F_k(v_k^t)$ as in Fig. \ref{fig:update_process}(a), the server-side models and the client-side models/classifiers are aggregated according to Fig. \ref{fig:update_process}(b).} 
      \label{fig:update_process}
      \vspace{-4mm}
\end{figure}

\textbf{Client-specific model aggregation.} For client $k$, $\phi_k$ combined with $h_k$ should work well on its local classes (personalized task), while  $\phi_k$ combined with $\theta$  should work well on all classes in the system. While the updated $\phi_k$ using (\ref{eq:sum_loss}) enables the client-side model ($\phi_k$ combined with $h_k$) to have strong personalization capability, it does not guarantee the generalization performance of the full model ($\phi_k$ combined with $\theta_k$). In particular,  since $\phi_k$ is updated only with the local data $D_k$ of client $k$, the output of $\phi_k$ (which becomes the input of $\theta$ in the full model) does not provide meaningful output features for the classes outside of $D_k$. 

A natural way to resolve this issue would be to aggregate  $\phi_k^{t+1}$ for all $k$ as $ \sum_{i=1}^K\alpha_i\phi_i^{t+1}$, and deploy this aggregated model at each client. However, this  can reduce the personalization capability at each client. In order to capture personalization while providing a meaningful result to the server-side model $\theta$, in SplitGP, each client $k$ computes weighted sum of $\phi_k^{t+1}$ and the average of $\phi_k^{t+1}$  for all $k=1,2,\dots,K$, as follows:
\vspace{-2mm}
\begin{equation}\label{eq:client_agg}
\phi_k^{t+1} \leftarrow \lambda \phi_k^{t+1} + (1-\lambda) \sum_{i=1}^K\alpha_i\phi_i^{t+1}.
\end{equation}
Here, $\lambda\in [0,1]$  controls the weights for personalization and generalization.  If  $\lambda=1$, the client-side model has a strong personalization capability but does not provide a meaningful output feature to the server-side model. If $\lambda=0$, the client-side model provides a generalizable feature to the server but lacks personalization capability.  
Using $\lambda$, the auxiliary classifiers $\{h_k\}_{k=1}^K$  are also aggregated at each client $k$ according to 
\vspace{-2mm}
\begin{equation}\label{eq:client_agg_classifier}
h_k^{t+1} \leftarrow \lambda h_k^{t+1} + (1-\lambda) \sum_{i=1}^K\alpha_ih_i^{t+1},
\end{equation}
which enables the client-side model to make reliable predictions on the out-of-distribution classes. Although generalization is not the main goal of the client model, conducting inference for out-of-distribution classes at the client when possible will further reduce communication cost and latency. Moreover, during inference, the client does not automatically know whether a datapoint is from one of its main classes or not. We therefore introduce a confidence threshold for the inference stage in Sec. III-C which chooses between client and server-side inference.

   Fig. \ref{fig:aggregation} summarizes the model aggregation step of our scheme. Note that, for simplicity of presentation, we have presented the model updates in (\ref{eq:phi_update}), (\ref{eq:h_update}),   (\ref{eq:theta_update}) assuming a single gradient step at each time $t$. In practice, these can be repeated multiple times in-between each  model aggregation process. 

After repeating the overall  process for $T$ global rounds, we obtain $K$ different personalized  models $\{\phi_k^T\}_{k=1}^K$ and classifiers $\{h_k^T\}_{k=1}^K$, and one server model $\theta^T$.  $\phi_k^T$ and $h_k^T$ are deployed at client $k$ while $\theta^T$ is implemented at the edge server. 

	\begin{algorithm}[t]
	\small
	\caption{SplitGP:  Training and Inference }\label{algo:proposed_final}
				\textbf{Training Phase}
		\begin{algorithmic}[1]
		\STATE \textbf{Input: }Initialized models $v^0=[\phi^0, h^0, \theta^0]$ \\
	\STATE \textbf{Output: } $v_k^T=[\phi_k^T, h_k^T, \theta^T]$ for each client $k=1,2,\dots,K$\\
		\FOR{each global round $t=0,1,\dots,T-1$}
		    \FOR{$k\in\{1,2,\dots,K\}$ \textbf{in parallel}}	
				    				  \STATE   $\ell_{C,k}(v_k^t)=  \frac{1}{|\tilde{D}_k^t|}\sum_{x\in \tilde{D}_k^t}\ell(x; \phi_k^t,h_k^t)$  \ \ \ // Client-side loss
				    \STATE  $ \ell_{S,k}(v_k^t)= \frac{1}{|\tilde{D}_k^t|}\sum_{x\in \tilde{D}_k^t}\ell(x; \phi_k^r,\theta_k^t)$  \ \ \ // Server-side loss
				    \STATE $F_k(v_k^t) = \gamma \ell_{C,k}(v_k^t) + (1-\gamma) \ell_{S,k}(v_k^t)
$  // Multi-exit loss 
    \STATE  $\phi_k^{t+1}  =   \phi_k^{t}  -  \eta_r\tilde{\nabla}_{\phi} F_k(v_k^t)$, 
$h_k^{t+1}  =   h_k^{t}  -  \eta_r\tilde{\nabla} F_k(v_k^t)$, $\theta_k^{t+1}  =   \theta_k^{t}  -  \eta_r\tilde{\nabla}_{\theta} F_k(v_k^t)$ \ \ \ // Model update
		    \ENDFOR
		    \STATE  $\theta^{t+1} \leftarrow\sum_{i=1}^K\alpha_i \theta_i^{t+1}$ \ \ // Server model aggregation
		    \STATE $\phi_k^{t+1} \leftarrow \lambda \phi_k^{t+1} + (1-\lambda) \sum_{i=1}^K\alpha_i \phi_i^{t+1}$   
		    		    \STATE $h_k^{t+1} \leftarrow \lambda \phi_k^{t+1} + (1-\lambda)\sum_{i=1}^K\alpha_i h_i^{t+1}$ // Client model aggregations; $\lambda$ controls the weights for personalization/generalization
		    \ENDFOR
		    		    \STATE $v_k^{T} = [\phi_{k}^{T}, h_{k}^{T}, \theta^{T}]$		
	\end{algorithmic}

		\textbf{Inference Phase}
		\begin{algorithmic}[1]
		\STATE \textbf{Input:} Test sample $z$ at client $k$ with $v_k^{T} = [\phi_{k}^{T}, h_{k}^{T}, \theta^{T}]$	
		\STATE \textbf{Output:} Prediction result for test sample $z$ \\
		    \STATE $E_k(z) = -\sum_{q=1}^Qp^{(q)}_k(z)\log  p^{(q)}_k(z)$  
			\IF{$E_k(z) < E_{th}$ }
			\STATE Make prediction with $\phi_k^T$ combined with $h_k^T$
			\ELSE 
			\STATE Make prediction with $\phi_k^T$ combined with $\theta^T$
			\ENDIF
	\end{algorithmic}	
\end{algorithm}
\subsection{Client-Side and Server-Side  Inference}
 During inference, each client $k$ must determine whether to rely on the client-side ($\phi_k^T, h_k^T$) or server-side ($\phi_k^T, \theta^T$) model. Given a test sample $z$ at client $k$,  the Shannon entropy is first computed using the client-side model ($\phi_k^T, h_k^T$) as 
$E_k(z) = -\sum_{q=1}^Qp^{(q)}_k(z)\log  p^{(q)}_k(z),$
where $Q$ is the total number of classes  in the system and $p^{(q)}_k(z)$ is the softmax output for class $q$ on sample $z$, using the  model deployed at client $k$.   If 
\vspace{-1mm}
\begin{equation}\label{eq:Eth}
E_k(z) \leq E_{th}
\end{equation}
holds for a desired entropy threshold $E_{th}$,   the inference is made at the client-side. Otherwise, i.e., if $E_k(z) > E_{th}$, the output feature of $\phi_k^T$ computed on sample $z$ is sent to the server and  the output of the server model $\theta^T$  is used for inference. 
The value of $E_{th}$ in (\ref{eq:Eth}) is therefore a control parameter for the amount of communication over the network during inference, while $\lambda$ in (\ref{eq:client_agg}) controls the weights for personalization and generalization. 
 We will analyze the effects of $E_{th}$ and $\lambda$ on SplitGP in Sec. VI.
The overall training process and inference stage of our scheme is described in Algorithm \ref{algo:proposed_final}.  
 



\section{Convergence Analysis}\label{sec:convergence}
We analyze the convergence behavior of  SplitGP based on some standard assumptions  in FL  \cite{li2019convergence, reisizadeh2020fedpaq, basu2019qsparse}.

\begin{assumption}
For each   $k$, $F_k(v)$ is $L$-smooth, i.e., $\|\nabla F_k(u)-\nabla F_k(v)\|\leq L\|u-v\|$ for any $u$ and $v$.
\end{assumption}
\begin{assumption}\label{assum:2}
For each   $k$, the expected squared norm of stochastic gradient is bounded, i.e., $\mathbb{E}[\|\tilde{\nabla}F_k(v)\|^2]\leq G$.
\end{assumption}
\begin{assumption}\label{assum:3}
The variance of  the stochastic gradient of $D_k$ is bounded, i.e., $\mathbb{E}[\|\nabla F_k(v) - \tilde{\nabla}F_k(v)\|^2]\leq \sigma_k^2$.
\end{assumption}
We also define the global loss function $F(v)$ as 
\vspace{-1mm}
\begin{equation}\label{eq:global_loss}
F(v)=\frac{1}{K}\sum_{k=1}^K  F_k(v),
\end{equation}
which is the   average of the losses defined in (\ref{eq:sum_loss}).  We show that our algorithm converges to a  stationary point of (\ref{eq:global_loss}), which guarantees the generalization capability of SplitGP while including personalization through $\lambda$ for any non-convex ML  loss function $F(v)$.
\subsection{Main Theorem and Discussions}
The following theorem gives   the convergence behavior of  SplitGP. 
The proof is given in Sec. \ref{subsec:proof}.
\begin{theorem}\label{thm:convergence}
\textbf{(SplitGP Convergence)} Let $\eta_t=\frac{\eta_0}{a+t}$, where $a=\frac{c+4}{1-\lambda^2}$ for some constant $c>0$. Suppose that $\eta_0$ is chosen to satisfy $\eta_t\leq \frac{1}{2L}$.  SplitGP model training converges as 
\vspace{-1mm}
\begin{align} \label{eq:thm_upperbound}
&\frac{1}{\Gamma_T}\sum_{t=0}^{T-1}\sum_{k=1}^K\frac{\eta_t}{4K}\mathbb{E}\Big[\|\nabla F(v_k^t)\|^2\Big]  \leq \frac{F(v^0) - F^*}{\Gamma_T}  \\
&+\frac{L\sum_{k=1}^K\sigma_k^2}{K}\left(\frac{1}{\Gamma_T} \sum_{t=0}^{T-1}\eta_t^2\right)  +  \epsilon(\lambda)\left(\frac{1}{\Gamma_T} \sum_{t=0}^{T-1}\eta_t^3\right),\nonumber 
\end{align}
where 
\begin{equation}\label{eq:ep_lambda}
\epsilon(\lambda) = \frac{16(c+4)G^2L^2\lambda^2(2-\lambda^2)}{c(1-\lambda^2)^2},
\end{equation}
$\Gamma_T = \sum_{t=0}^{T-1}\eta_t$ and $F^*$ is the minimum value of   $F(v)$    in (\ref{eq:global_loss}).
\end{theorem}
Here,  $\epsilon(\lambda)$ is the term   specific to our work, arising from the joint consideration of generalization and personalization. 
By setting  $\eta_t=\frac{\eta_0}{a+t}$, we have  $\Gamma_T = \sum_{t=0}^{T-1}\eta_t\rightarrow \infty$ as $T$ grows, and  $\sum_{t=0}^{\infty}\eta_t^2<\infty$,  $\sum_{t=0}^{\infty}\eta_t^3<\infty$. Hence, for any $\lambda \in [0,1)$, the upper bound in (\ref{eq:thm_upperbound}) goes to 0 as $T$ grows. Thus,   we have $\underset{t\in\{0,1,\dots, T-1\}}{\text{min}} \mathbb{E}[\|\nabla F(v_k^t)\|]\overset{T\rightarrow \infty}{\longrightarrow}0$ for all $k=1,..., K$, which guarantees  convergence  to a stationary point of (\ref{eq:global_loss}). 

Theorem \ref{thm:convergence} indicates that $v_k^t=[\phi_k^t, h_k^t, \theta^t]$, which has a certain amount of  personalization capability from $\lambda$, also obtains the generalization capability of (\ref{eq:global_loss}). In other words, both personalization and generalization are achieved. 
Here, as $\lambda$ grows,   a larger number of global rounds is required to reduce the upper bound in (\ref{eq:thm_upperbound}); this is the cost for achieving a stronger personalization  at the client-side. Note that the case with $\lambda=1$ does not guarantee convergence, since the client-side models are not aggregated. On the other hand, the case with $\lambda=0$ reduces to the bound of conventional FL. 

\subsection{Convergence Proof}\label{subsec:proof}
Using $v_k^t=[\phi_k^t, h_k^t, \theta^t]$, we first define $v^t$ as:
\vspace{-0.7mm}
\begin{align} 
v^t=\frac{1}{K}\sum_{k=1}^Kv_k^t. 
\end{align}
By  the $L$-smoothness of $F(v)$ and taking  the expectation of  both sides, we have
\vspace{-1mm}
\begin{align}\label{eq:main_goal}
\mathbb{E}[F(v^{t+1})] - \mathbb{E}[F(v^t)] &\leq \underbrace{\mathbb{E}[\langle \nabla F(v^t), v^{t+1} - v^t\rangle]}_{A} \nonumber  \\ 
&+   \underbrace{\frac{L}{2}\mathbb{E}[\|v^{t+1}-v^t\|^2]}_{B}. 
\end{align} 

\textbf{Step 1: Bounding $A$.} We first rewrite $A$ as follows: 
\vspace{-0.5mm}
{\small
\begin{align}
& A  \underset{(a)}{=} -\eta_t \mathbb{E}\Big[\Big\langle \nabla F(v^t), \frac{1}{K}\sum_{k=1}^K\tilde{\nabla} F_k(v_k^t)\Big\rangle\Big] \\
& \underset{(b)}{=} -\eta_t \mathbb{E}\Big[\Big\langle \nabla F(v^t), \frac{1}{K}\sum_{k=1}^K\nabla F_k(v_k^t)\Big\rangle\Big]
  \underset{(c)}{=} \underbrace{-\frac{\eta_t}{2}  \mathbb{E}\Big[ \|\nabla F(v^t)\|^2\Big]}_{A_1}\nonumber\\  & -\frac{\eta_t}{2}  \mathbb{E}\Big[\Big\|\frac{1}{K}\sum_{k=1}^K\nabla F_k(v_k^t)\Big\|^2   - \underbrace{\Big\|\nabla F(v^t) -  \frac{1}{K}\sum_{k=1}^K \nabla F_k(v_k^t)\Big\|^2}_{A_2}\Big], \nonumber
\end{align} }
where $(a)$ comes from  $v^{t+1} - v^t = -\eta_t\frac{1}{K}\sum_{k=1}^K\tilde{\nabla} F_k(v_k^t)$, $(b)$ follows from taking the expectation for the mini-batch, and $(c)$ is obtained by utilizing $\|z_1-z_2\|^2 = \|z_1\|^2 + \|z_2\|^2 - 2\langle z_1, z_2\rangle$.

We now focus on $A_1$. We can write
\vspace{-0.5mm}
\begin{align}
&\|\nabla F(v^t)\|^2  \underset{(d)}{\geq} \frac{1}{2}\|\nabla F(v_k^t)\|^2 - \| \nabla F(v_k^t) - \nabla F(v^t)   \|^2 \nonumber\\
& = \frac{1}{2}\|\nabla F(v_k^t)\|^2 -\Big\|\frac{1}{K}\sum_{i=1}^K (\nabla F_i(v_k^t) - \nabla F_i(v^t))\Big\|^2 \nonumber\\
& \underset{(e)}{\geq} \frac{1}{2}\|\nabla F(v_k^t)\|^2 - L^2 \| v_k^t - v^t \|^2 
\end{align}
for any $k$. Here, $(d)$ comes from using  $\|a+b\|^2 \leq 2\|a\|^2 + 2\|b\|^2$ and $(e)$ comes from  $L$-smoothness.  Thus, we can bound $A_1$ as
\vspace{-0.5mm}
\begin{align}
&A_1 = -\frac{\eta_t}{2} \mathbb{E}\Big[\|\nabla F(v^t)\|^2\Big]
 = -\frac{\eta_t}{2K}\sum_{k=1}^K\mathbb{E}\Big[\|\nabla F(v^t)\|^2\Big]\nonumber \\
&\leq -\frac{\eta_t}{4K}\sum_{k=1}^K\mathbb{E}\Big[\|\nabla F(v_k^t)\|^2\Big] + \frac{\eta_tL^2}{2K}\sum_{k=1}^K \mathbb{E}[\| v_k^t - v^t \|^2 ].
\end{align}
For $A_2$, we have 
\vspace{-0.5mm}
\begin{align}
\frac{\eta_t}{2}\mathbb{E}[A_2]&=\frac{\eta_t}{2}\mathbb{E}\Big[\Big\|\frac{1}{K}\sum_{k=1}^K (\nabla F_k(v^t) - \nabla F_k(v_k^t))\Big\|^2\Big]\nonumber\\
&\underset{(f)}{\leq} 
 \frac{\eta_t }{2K}  \sum_{k=1}^K \mathbb{E}[ \|\nabla F_k(v^t) - \nabla F_k(v_k^t)\|^2] \nonumber\\ & 
 \underset{(g)}{\leq} \frac{\eta_tL^2 }{2K}  \sum_{k=1}^K \mathbb{E}[ \| v^t -  v_k^t \|^2], 
\end{align}
where $(f)$ holds due to the convexity of $\|\cdot\|^2$ and $(g)$ holds  due to the $L$-smoothness assumption.

\textbf{Step 2: Bounding $B$.} Now we bound the term $B$. By utilizing $v^{t+1} - v^t = -\eta_t\frac{1}{K}\sum_{k=1}^K\tilde{\nabla} F_k(v_k^t)$, we can write $B \leq \eta_t^2 L( \mathbb{E}[ \|\frac{1}{K}\sum_{k=1}^K\nabla F_k(v_k^t) \|^2] + \mathbb{E}[\|\frac{1}{K}\sum_{k=1}^K\nabla F_k(v_k^t) - \frac{1}{K}\sum_{k=1}^K\tilde{\nabla} F_k(v_k^t)\|^2] )$, where
\begin{align}
& \mathbb{E}\Big[\Big\|\frac{1}{K}\sum_{k=1}^K\nabla F_k(v_k^t) - \frac{1}{K}\sum_{k=1}^K\tilde{\nabla} F_k(v_k^t)\Big\|^2\Big] \\
&\leq \frac{1}{K} \sum_{k=1}^K  \mathbb{E}[\| \nabla F_k(v_k^t) -  \tilde{\nabla} F_k(v_k^t)\|^2] \Big)
 \underset{(h)}{\leq}    \frac{1}{K}\sum_{k=1}^K\sigma_k^2\nonumber
\end{align}
and $(h)$ results from Assumption \ref{assum:3}.

By inserting the bounds of $A$ and $B$ to (\ref{eq:main_goal}), and by employing a  learning rate that satisfies $\eta_t\leq \frac{1}{2L}$,  we obtain
\begin{align}\label{eq:goal_modified2}
\frac{\eta_t}{4K}\sum_{k=1}^K\mathbb{E}\Big[\|&\nabla F(v_k^t)\|^2\Big]  \leq \mathbb{E}[F(v^t)] - \mathbb{E}[F(v^{t+1})]\nonumber\\
 &+\frac{\eta_t^2L}{K}\sum_{k=1}^K\sigma_k^2 + \underbrace{\frac{\eta_tL^2 }{K} \sum_{k=1}^K \mathbb{E}[\| v_k^t - v^t \|^2 ]}_{C}. 
\end{align}

\textbf{Step 3: Bounding $C$.}  To bound $C$, we  define 
\begin{equation}
\psi_k^t = [\phi_k^t, h_k^t] 
\end{equation}
\begin{equation}
\hat{\psi}_k^{t+1} =  \psi_k^t - \eta_t \tilde{\nabla}_{\psi} F_k(v_k^t), 
\end{equation}
 $\psi^t=\frac{1}{K}\sum_{k=1}^K\psi_k^t$ and  $\hat{\psi}^t=\frac{1}{K}\sum_{k=1}^K\hat{\psi}_k^t$.  Then,  we have
\begin{align}
\psi_k^{t+1} = \lambda \hat{\psi}_k^{t+1} + (1-\lambda)\hat{\psi}^{t+1} \label{eq:property1}
\end{align}
 and   $\frac{1}{K}\sum_{k=1}^K\|v_k^t - v^t\|^2 =\frac{1}{K}\sum_{k=1}^K\|\psi_k^t - \psi^t\|^2$. We can  write
\begin{align}
&\frac{1}{K}\sum_{k=1}^K\|\psi_k^t - \psi^t\|^2 
= \frac{1}{K}\sum_{k=1}^K\|(\psi_k^t- \hat{\psi}^t) - (\psi^t- \hat{\psi}^t)\|^2\nonumber\\
&\underset{(i)}{\leq} \frac{1}{K}\sum_{k=1}^K\|\psi_k^t - \hat{\psi}^t\|^2
 \underset{(j)}{=} \frac{1}{K}\sum_{k=1}^K\|\lambda(\hat{\psi}_k^t - \hat{\psi}^t)\|^2\nonumber\\ \label{eq20}
&=\frac{\lambda^2}{K}\sum_{k=1}^K\|(\hat{\psi}_k^t - \psi^{t-1}) - (\hat{\psi}^t - \psi^{t-1})\|^2 \nonumber \\
&\underset{(k)}{\leq} \frac{\lambda^2}{K}\sum_{k=1}^K\|\hat{\psi}_k^t - \psi^{t-1}\|^2\nonumber\\
 & \underset{(l)}{=}  \frac{\lambda^2}{K}\sum_{k=1}^K\|-\eta_{t-1}\tilde{\nabla}_\psi F_k(v_k^{t-1}) + (\psi_k^{t-1} - \psi^{t-1})\|^2
 \end{align}
where $(i)$ and $(k)$ come from $\mathbb{E}[\|z - \mathbb{E}[z]\|^2]\leq \mathbb{E}[\|z\|^2]$, $(j)$ follows from  (\ref{eq:property1}) and $(l)$ results from  $\hat{\psi}_k^{t+1} = \psi_k^t - \eta_t \tilde{\nabla}_{\psi} F_k(v_k^t)$. 
Now following the proof of Lemma 4 of \cite{basu2019qsparse} and utilizing Assumption \ref{assum:2}, when $\eta_t = \frac{\eta_0}{a+t}$ and $a=\frac{c+4}{1-\lambda^2}$ for some constant $c>0$, we have 
$\frac{1}{K} \sum_{k=1}^K \mathbb{E}[\| v_k^t - v^t \|^2 ] \leq \frac{16(c+4)G^2\lambda^2(2-\lambda^2)}{c(1-\lambda^2)^2}$.

\textbf{Step 4: Telescoping sum.} Finally, after inserting the result of Step 3 into (\ref{eq:goal_modified2}), we have
\begin{align}
\frac{\eta_t}{4K}\sum_{k=1}^K&\mathbb{E}\Big[\|\nabla F(v_k^t)\|^2\Big]  \leq \mathbb{E}[F(v^t)] - \mathbb{E}[F(v^{t+1})]\nonumber\\ &+\frac{\eta_t^2L}{K}\sum_{k=1}^K\sigma_k^2 + \frac{16\eta_t^3(c+4)G^2\lambda^2(2-\lambda^2)}{c(1-\lambda^2)^2}.
\end{align}
After summing up for $t=0,1\dots,T-1$ and dividing both sides by $\Gamma_T=\sum_{t=0}^{T-1}\eta_t$, with some manipulations, we obtain (\ref{eq:thm_upperbound}). This completes the proof of Theorem \ref{thm:convergence}. 


   \begin{table*}[ht]	
	\caption{\small \textbf{Resources required and latency incurred  at each client during inference:} comparing SplitGP with baselines.}
	\centering
	\label{table:cost_compare}
	\begin{tabular}{l|c|c|c|c}
		\toprule  
		   \multicolumn{1}{c}{Methods} & \multicolumn{1}{c}{Storage} & \multicolumn{1}{c}{Computation} & \multicolumn{1}{c}{Communication}  & \multicolumn{1}{c}{Inference time} \\
		   		\midrule
						Full model  at the server-side  & $0$ & $0$ & $q|D|$& $\frac{q|D|}{R} + \frac{(|\phi|+|\theta|)|D|}{P_S}$	  \\	

		Full model  at the client-side & $|\phi|+|\theta|$ & $(|\phi|+|\theta|)|D|$ & $0$ & $\frac{(|\phi|+|\theta|)|D|}{P_C}$	  \\	
		Proposed framework (SplitGP)    & $|\phi|+|h|$ & $(|\phi|+|h|)|D|$& $\beta q_c|D|$ & $\frac{(|\phi|+|h|)|D|}{P_C}+\frac{\beta q_c|D|}{R} + \frac{\beta|\theta||D|}{P_S}$ \\	
		\bottomrule
	\end{tabular}
	\vspace{-3mm}
\end{table*}
\section{Inference-Time Analysis and Model Splitting}\label{sec:latency}

In this section, we analyze  the storage, computation, communication and time  required during the inference stage. 
\subsection{Notations and Assumptions}
Let $P_C$ and $P_S$ be the available  computing powers of each client and the server, respectively. Let $|\phi|$, $|\theta|$, $|h|$ be the numbers of parameters of $\phi$, $\theta$, $h$, respectively. 
Considering  $P_C \ll P_S$ in practice, we split the model $w=[\phi, \theta]$ such that the size of client-side component $\phi$ is significantly smaller than the server-side component $\theta$, i.e.,  $|\phi|\ll|\theta|$.        Moreover, $h$ is assumed to be a small classifier satisfying $|h|\ll|\phi|$ and $|h|\ll|\theta|$.  To make  our analysis tractable, we assume that the inference time (i.e., time required for forward propagation through the neural network) is proportional to the number of   parameters of the model \cite{han2021accelerating, canziani2016analysis}.  For example,  given a model with $|\phi|+|h|$ parameters and a   test dataset of size $|D|$, the inference time at the client will be proportional to $\frac{(|\phi|+|h|)|D|}{P_C}$. One may also consider different latency models  which is out of scope of this paper. We   define $R$ as the uplink data rate between a single client and the server. $\beta$ is the portion of total test samples that are inferred at the client-side as a result of (\ref{eq:Eth}), while $1-\beta$ is the portion of test samples inferred at the server-side. Finally, $q_c$ denotes the dimension of the cut-layer (i.e., output dimension of $\phi$)  and $q$ denotes the size of the test sample (input dimension of the model). We assume the same $q_c$ for all client layers for analytical tractability. 

\subsection{Resource and Latency Analysis}

Table \ref{table:cost_compare} compares   our methodology  with existing frameworks during inference.  We present the derivations in the following:

\textbf{Full model at the client.} When the full model $w=[\phi, \theta]$ is implemented at individual clients, 
the required storage   for each client  is $|\phi|+|\theta|$. Hence, the client-side computational load becomes $(|\phi|+|\theta|)|D|$. Since all predictions are made at the client, no communication is required during inference. Hence, the inference time can be written as follows:
\vspace{-1mm}
\begin{equation}\label{eq:client_latency}
\tau_1= \frac{(|\phi|+|\theta|)|D|}{P_C}.
\end{equation}

\textbf{Full model at the server.}  When the full model is deployed at the edge server, client-side storage is unused  and the client-side computational load is also zero. Since the entire test set must be sent to the server,    the required communication load during inference becomes $q|D|$. The inference time can be written as the sum of communication time and server-side computation time as follows:
\vspace{-1.5mm}
\begin{equation}\label{eq:server_latency}
\tau_2= \frac{q|D|}{R} + \frac{(|\phi|+|\theta|)|D|}{P_S}. 
\end{equation}

\textbf{Proposed SplitGP.} In our approach, the required storage space at each client is  $|\phi|+|h|$ while   the  server-side storage is   $|\theta|$. The client-side computation is written as $(|\phi|+ |h|)|D|$.  Given the cut-layer dimension $q_c$,  $\beta$ portion of test samples are predicted at the server-side, requiring a   communication load of $\beta q_c|D|$.  For inference time, given  a test sample $z$, forward propagation is first performed at the client-side, which has a latency of $\frac{|\phi|+|h|}{P_C}$. If $E_t(k)\leq E_{th}$ (with probability $1-\beta$), the prediction is made at the client  which  requires no additional time. Otherwise,   with probability $\beta$, each client sends the output feature of $\phi_k$ to the server, which requires an additional latency of $\frac{q_c}{R}$ for communication and $\frac{|\theta|}{P_S}$ for server-side computation. Hence,  the latency of SplitGP is written as
\vspace{-1.5mm}
\begin{equation}\label{eq:proposed_latency}
\tau = \frac{(|\phi|+|h|)|D|}{P_C}+\frac{\beta q_c|D|}{R} + \frac{\beta|\theta||D|}{P_S}. 
\end{equation}

Based on this analysis, we pose the following two questions: (i) How should we split the model $w$ into $\phi$ and $\theta$ in practice? (ii) When is our framework with split models beneficial compared to other baselines  in terms of inference time?

\subsection{Model Splitting and Feasible Regimes}
For the first question above, note that model splitting gives a trade-off between  client-side personalization capability and  inference time: as we increase the size  of the client-side component $\phi$,   personalization    improves but the inference time increases according to (\ref{eq:proposed_latency}). Let $|\phi_{\text{min}}|$ be the minimum size of $\phi$ to achieve a desired level of personalization capability at the client, which   must be selected empirically based on the local dataset and observed ML task difficulty.     We also let $\tau'$ be the latency per test sample that the system should support. 
Based on these constraints, we  state the following result: 
\vspace{-1mm}
\begin{proposition}
From the  latency constraint $\frac{\tau}{|D|}\leq \tau'$ and  the personalization constraint $|\phi|\geq|\phi_{\text{min}}|$,  the  feasible model splitting regime  for the client-side component is given as
\begin{equation}\label{eq:phi_range} 
|\phi_{\text{min}}| \leq  |\phi| \leq \frac{P_CP_S( \tau'  - \frac{\beta q_c}{R} - \frac{|h|}{P_C}  - \frac{\beta|w|}{P_S})}{P_S-\beta P_C }.
\end{equation}
\end{proposition}
Note that $P_S-\beta P_C >0$ holds since $P_C<P_S$ and $\beta\leq 1$.
 When improving accuracy prioritized over improving latency, we can split the model   to satisfy $|\phi| = \frac{P_CP_S( \tau'  - \frac{\beta q_c}{R} - \frac{|h|}{P_C}   - \frac{\beta|w|}{P_S})}{P_S-\beta P_C }$, i.e., increase the size of the client-side component as much as possible.   When latency is prioritized, we choose  $|\phi| = \phi_{\text{min}}$, to minimize the inference time while achieving the minimum required personalization capability at the client-side.  



Now we turn to the second question, considering a fixed model splitting $w=[\phi, \theta]$.   We first compare with the case where the full model is deployed at the client. From  (\ref{eq:client_latency}) and (\ref{eq:proposed_latency}), we have the following proposition: 
\vspace{-1mm}
\begin{proposition}
We have $\tau\leq \tau_1$   if and only if 
\begin{equation}\label{eq:thm1}
P_C\leq \frac{|\theta|-|h|}{\beta(\frac{q_c}{R} + \frac{|\theta|}{P_S})}.
\end{equation}
\end{proposition}
The above result indicates that  our solution is faster than the  baseline   when the client-side computing power $P_C$ is smaller than a specific threshold. This makes intuitive  sense because deploying the full model at the client-side incur significant inference latency when  $P_C$ is  small (e.g., low-cost IoT devices). 


 
 \begin{figure}
         \vspace{-1mm}
\centering
  \subfigure[Inference time versus $P_C$]{\includegraphics[width=0.24\textwidth]{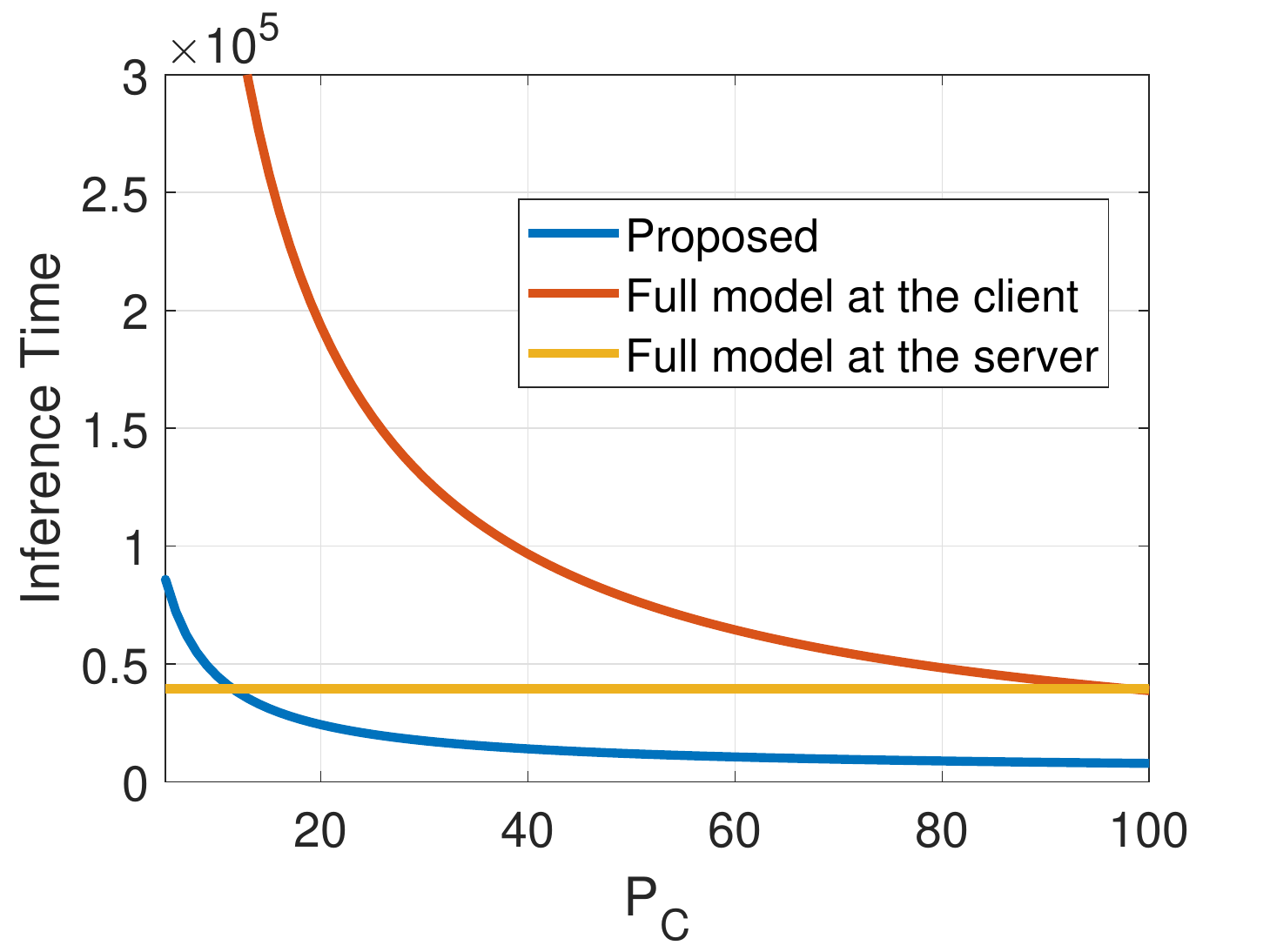}\label{fig:P_C}}
      \subfigure[Inference time versus $R$]{\includegraphics[width =0.24\textwidth]{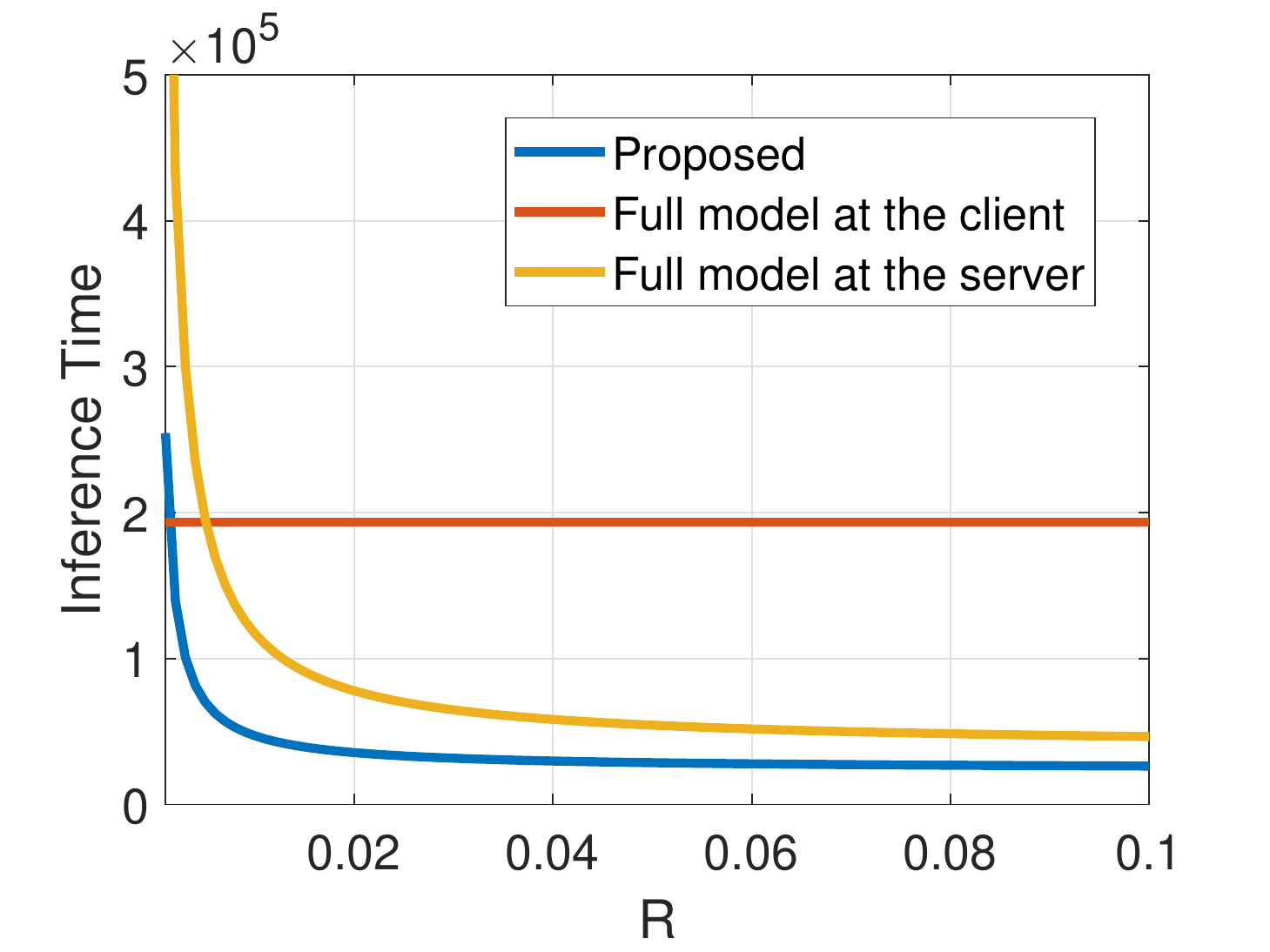}\label{fig:R}}
        \vspace{-2mm}
  \caption{ \small  Inference time depending on client-side computing power $P_C$ and communication rate $R$. Our framework   demonstrates significant advantage compared to the baselines   in  various  power-rate regimes. }
  \vspace{-4mm}
    \label{fig:latency}
\end{figure}


Second, we compare with the baseline where the full model is implemented at the edge server during inference. Based on (\ref{eq:server_latency}) and (\ref{eq:proposed_latency}), we state the following proposition: 
\vspace{-1mm}
\begin{proposition}
We have $\tau\leq \tau_2$    if and only if 
\begin{equation}\label{eq:server_beneficial}
R\leq \frac{q-\beta q_c}{\frac{|\phi|+|h|}{P_C} - \frac{(1-\beta)(|\phi|+|\theta|)}{P_S}}.
\end{equation}
\end{proposition}
According to (\ref{eq:server_beneficial}), our solution is beneficial when the communication rate $R$ is smaller than a specific value, since this baseline requires transmission of all  test samples from   client to  server.   

Fig. \ref{fig:latency} shows the inference times of the models in Table \ref{table:cost_compare} with $|\phi| = 387,840$, $|\theta|=3,480,330$, $|h|=23,050$, which corresponds  to the convolutional neural network (CNN) that is utilized for experiments in the next section. Other parameters are $P_S=100$, $P_C=20$, $R=1$, $\beta=0.1$, $|D|=1$. It can be seen that our framework achieves smaller inference time compared to existing baselines in various  $P_C$ and $R$ regimes. 




\section{Experimental Results} \label{sec:exp}
We evaluate our method on Fashion-MNIST (FMNIST) \cite{xiao2017fashion}  and CIFAR-10 \cite{krizhevsky2009learning}. Both datasets contain $10$ classes. We utilize a CNN   with 5 convolutional layers and 3 fully connected layers for FMNIST dataset. For CIFAR-10, we adopt VGG-11. 

\textbf{Implementation.} We consider   $K=50$ clients.   To model  non-IID data distributions, following the setup of \cite{mcmahan2017communication}, we first sort the overall train set based on classes and divide it into 100 shards. 
We then randomly allocate 2 shards to each client.  We used a  learning rate of $\eta=0.01$ for all schemes. In each global round, each  client updates its model for one epoch with a mini-batch size of 50, and cross-entropy loss is utilized throughout the training process. Moreover, we set $\lambda=0.2$  and choose the optimal $E_{th}\in\{0.05, 0.1, 0.2, 0.4, 0.8, 1.2, 1.6,  2.3\}$ unless otherwise stated.  We train the CNN model with FMNIST for 120 global rounds and VGG-11 model with CIFAR-10 for 800 global rounds. For our scheme, we  split the full CNN model  (for FMNIST) such that  the client-side  $\phi$ contains 4 convolutional layers ($|\phi|=387,840$) and  the server-side  $\theta$ contains 1 convolutional layer and 3 fully connected layers ($|\theta|=3,480,330$). The fully connected layer with size $|h|=23,050$ is utilized as the auxiliary classifier. We also split the VGG-11   as $|\phi|=972,554$ and $|\theta|=8,258,560$, and adopt the fully connected layer with size  $|h|=10,250$ as a classifier. 

\textbf{Baselines.} We compare  SplitGP with the following baselines. First, we consider the personalized FL scheme proposed in \cite{deng2020adaptive}, where the trained personalized models are deployed at individual clients during inference.  We also consider a generalized global model constructed via conventional  FL \cite{mcmahan2017communication} as well as SplitFed \cite{thapa2022splitfed}. Note that FL and SplitFed produce the same model while SplitFed can save   storage and computation resources during training. This generalized global model can be deployed either at the client or at the server.   Finally, we consider a multi-exit neural network that has two exits, one at the client-side and the other at the server-side,  constructed via FL or SL. 
 For a fair comparison, FedAvg \cite{mcmahan2017communication} is adopted for the model aggregation process of all schemes. 

\textbf{Evaluation.} When training is finished,  the overall performance  is measured by averaging the local test  accuracies of all clients. 
We construct the local test set of each client as the union of the main test samples and the out-of-distribution test samples. The main test samples are  constructed by selecting all test samples of the main classes, e.g., if client $k$ has only classes $1$ and $2$ in its local data,   all the test samples with classes $1$ and $2$ in the original test set are selected to construct  the main test samples. When constructing the out-of-distribution test samples, we utilize the relative portion of out-of-distribution test samples $\rho$ defined in (\ref{eq:rho}). 
Given the main test samples,  a fraction $\rho$ of out-of-distribution  samples are selected  from the original test  set. We reiterate that the previous works on personalized FL  adopted  $\rho=0$ for evaluation. 
\begin{figure}[t]
\vspace{-2mm}
\centering
  \subfigure[FMNIST]{\includegraphics[width=0.24\textwidth]{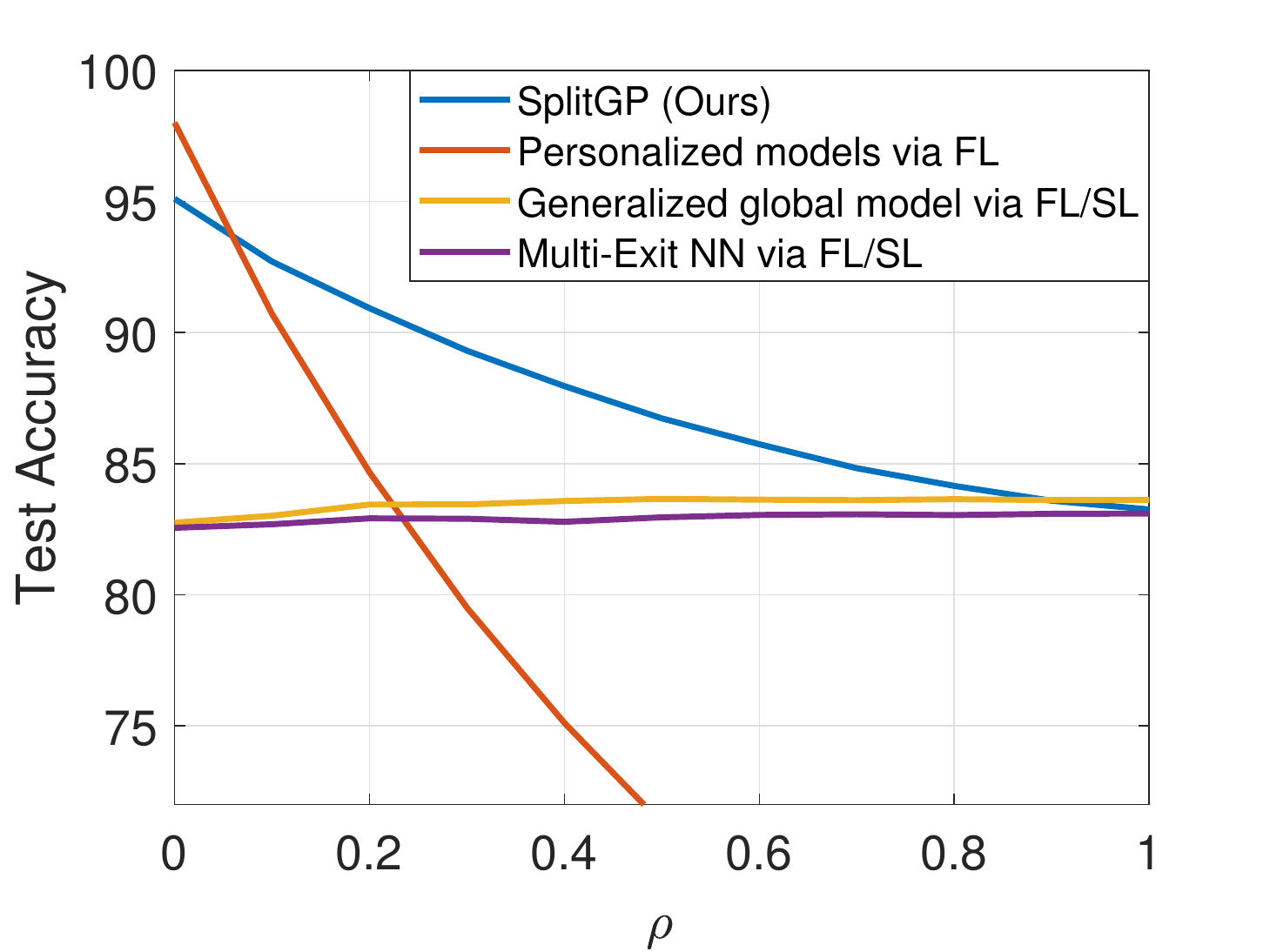}}
  \subfigure[CIFAR-10]{\includegraphics[width =0.24\textwidth]{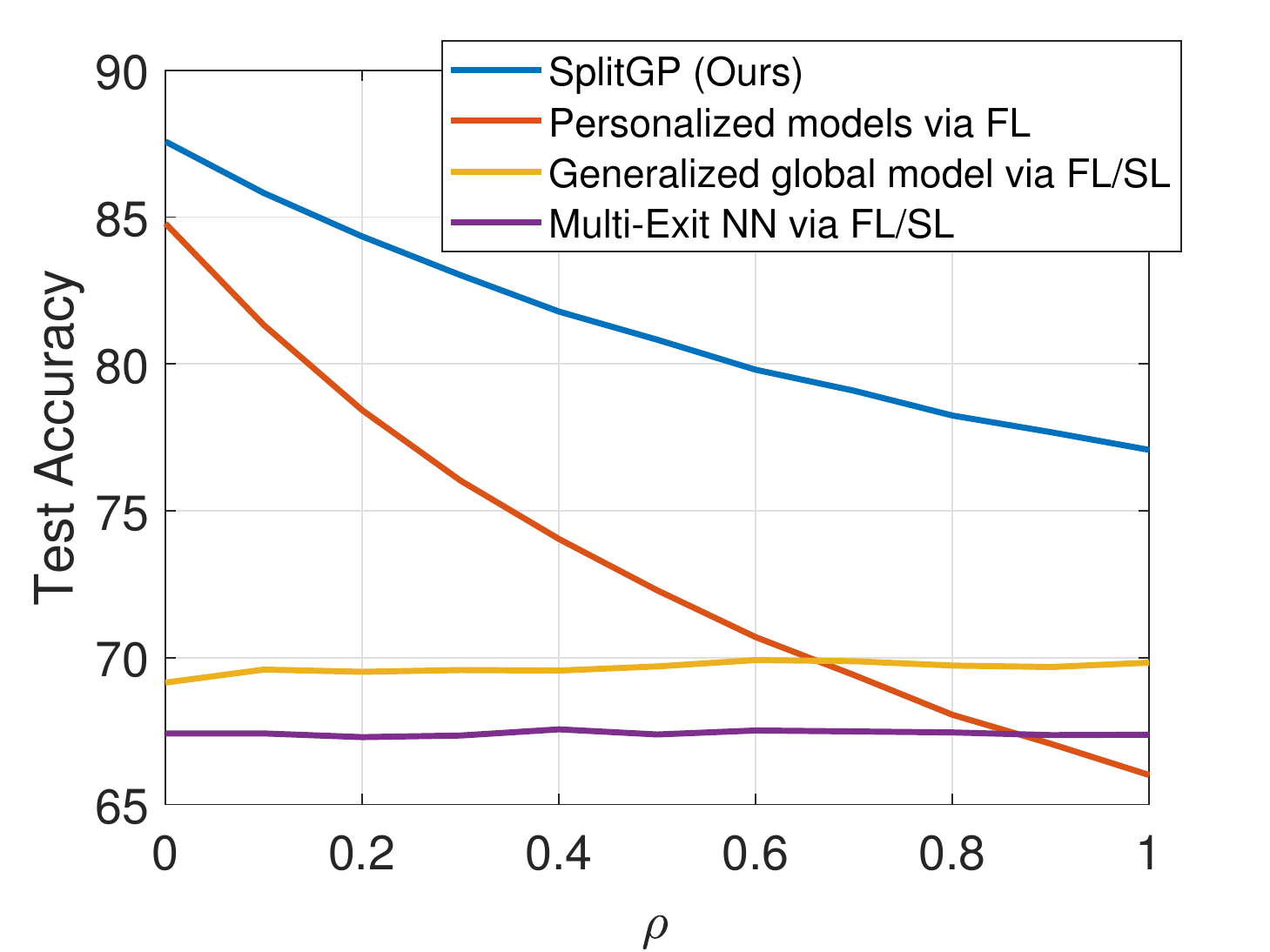}}
    \vspace{-3mm}
  \caption{\small Test accuracy vs. $\rho$. By capturing both personalization and generalization, SplitGP has   advantages for most settings of $\rho$. } 
    \label{fig:accuracy_vs_ratio}
        \vspace{-1mm}
\end{figure}
\begin{table}
\scriptsize
	\caption{\small  Effect of out-of-distribution test samples on FMNIST.  }
	\vspace{-1mm}
	\centering
	\label{table:main_1}
	\begin{tabular}{l||ccccc}
		\toprule  
		\textbf{Methods} &  $\rho=0$ & $\rho=0.2$ &$\rho=0.4$ &$\rho=0.6$ &$\rho=0.8$ \\ 
		\midrule
		Personalized FL&  $\mathbf{98.00}\%$& $84.67\%$&   $75.11\%$ & $67.96\%$ & $62.43\%$  \\ 
		Generalized FL& $82.75\%$& $83.44\%$&   $83.57\%$ & $83.62\%$ & $83.64\%$  \\ 
	SplitGP (Ours)&  $95.10\%$& $\mathbf{90.93}\%$  &   $\mathbf{87.95}\%$ &$\mathbf{85.74}\%$ & $\mathbf{84.15}\%$ \\ 
		\bottomrule
	\end{tabular}
	  \vspace{-3mm}
\end{table}

\begin{figure*}
    \vspace{-2mm}
\centering
  \subfigure[FMNIST, $\rho=0.2$]{\includegraphics[width=0.24\textwidth]{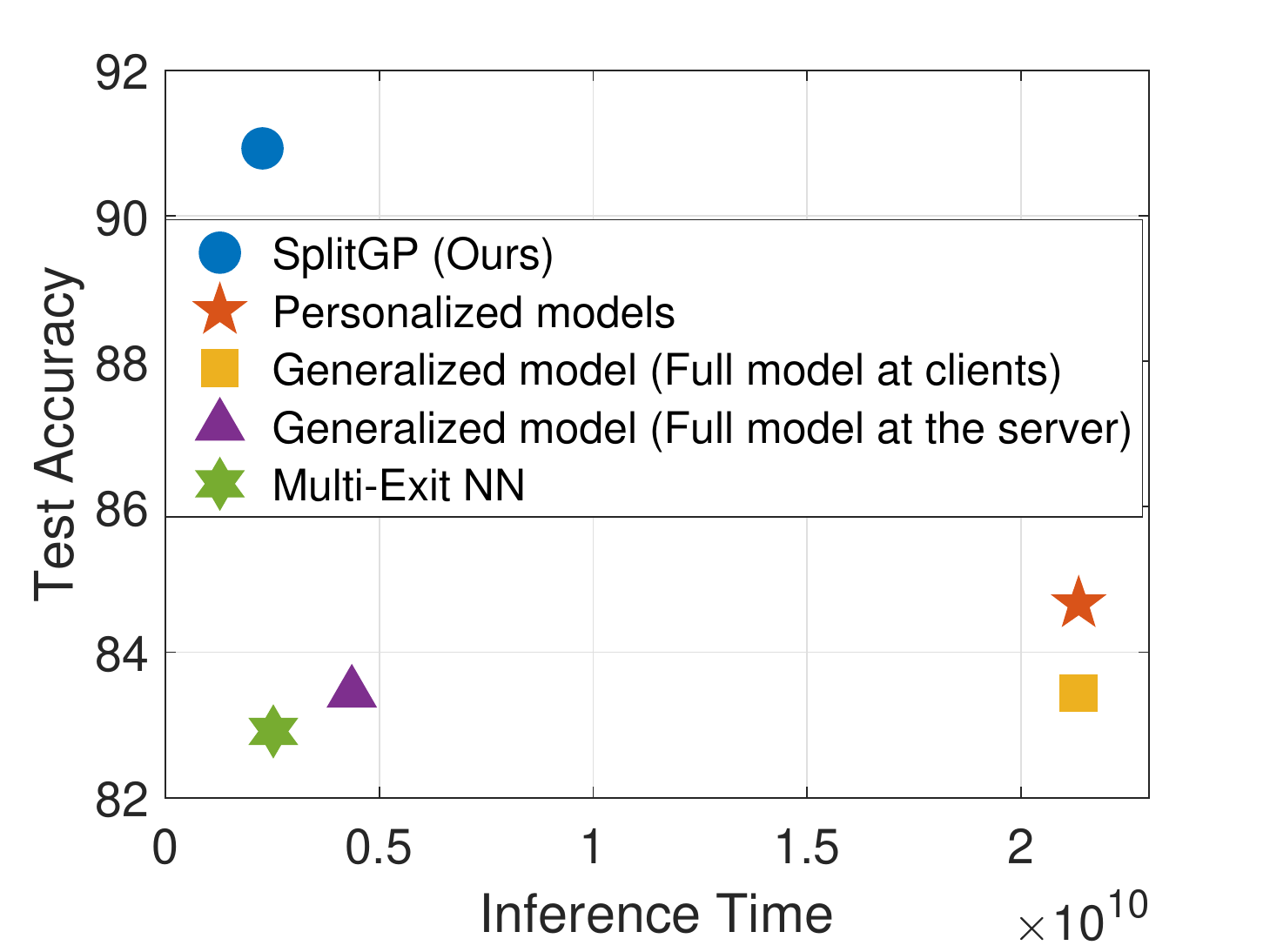}}
    \vspace{-1.7mm}
  \subfigure[FMNIST, $\rho=0.4$]{\includegraphics[width=0.24\textwidth]{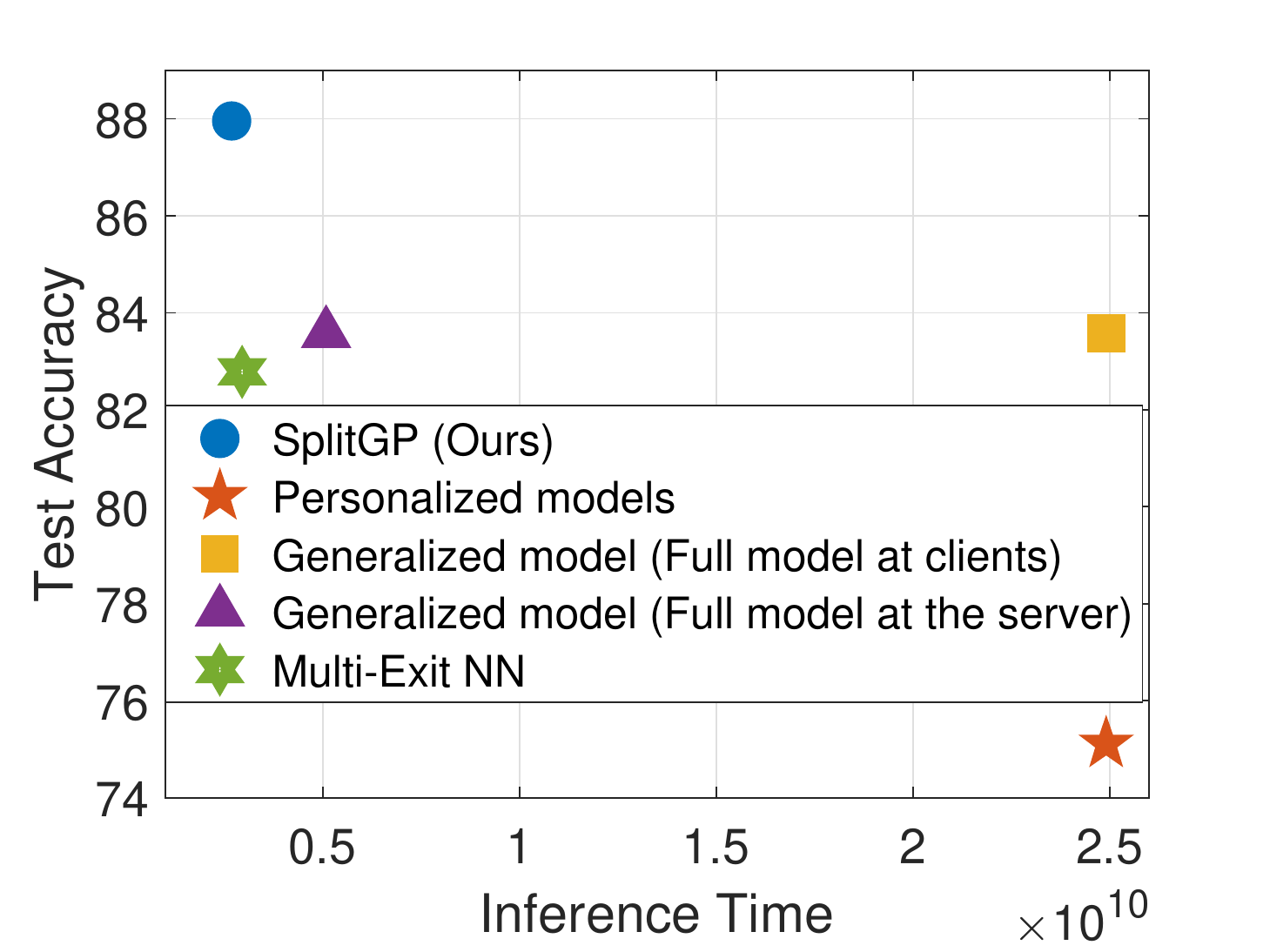}}
        \subfigure[FMNIST, $\rho=0.6$]{\includegraphics[width =0.24\textwidth]{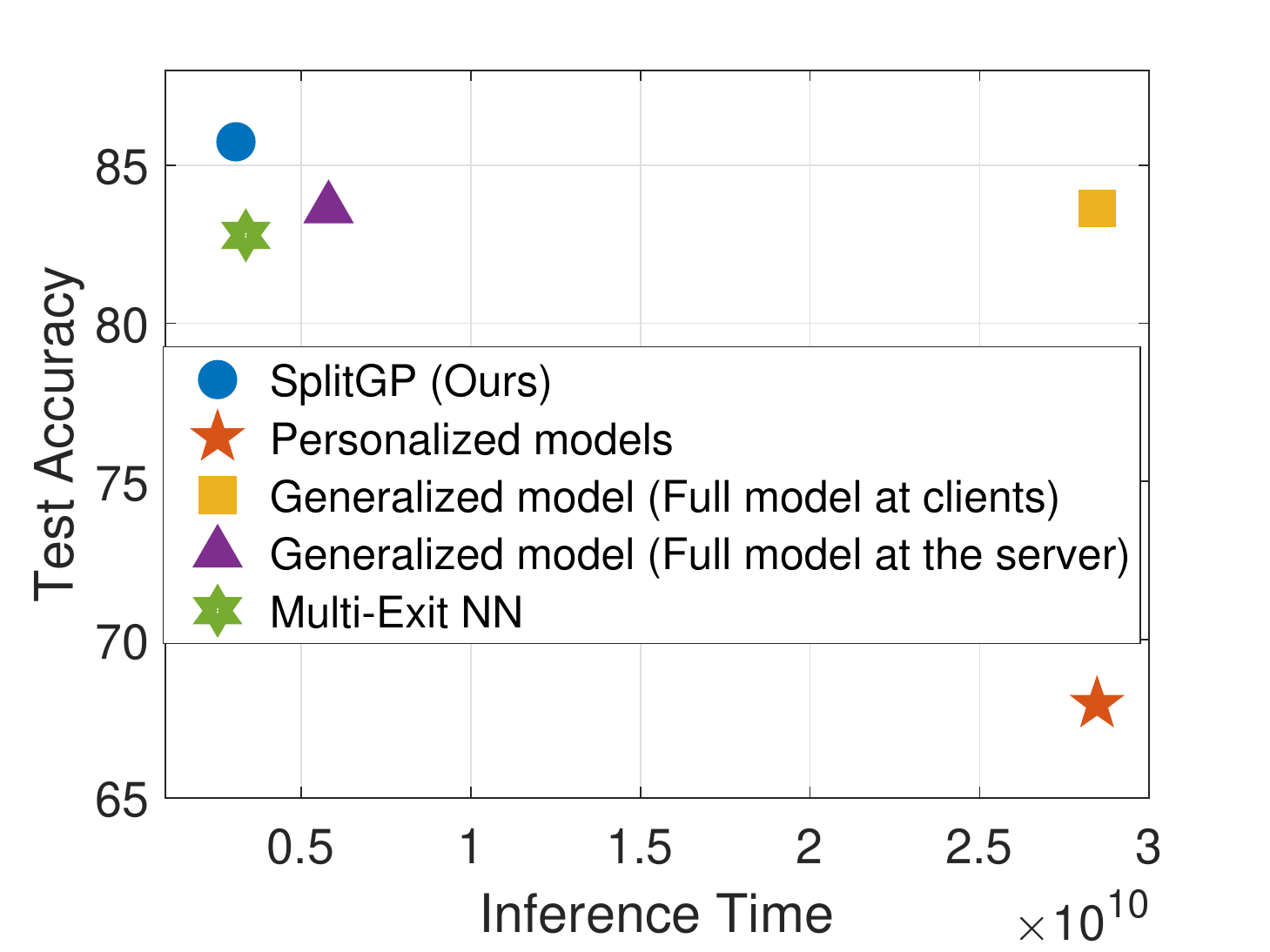}}
      \subfigure[FMNIST, $\rho=0.8$]{\includegraphics[width =0.24\textwidth]{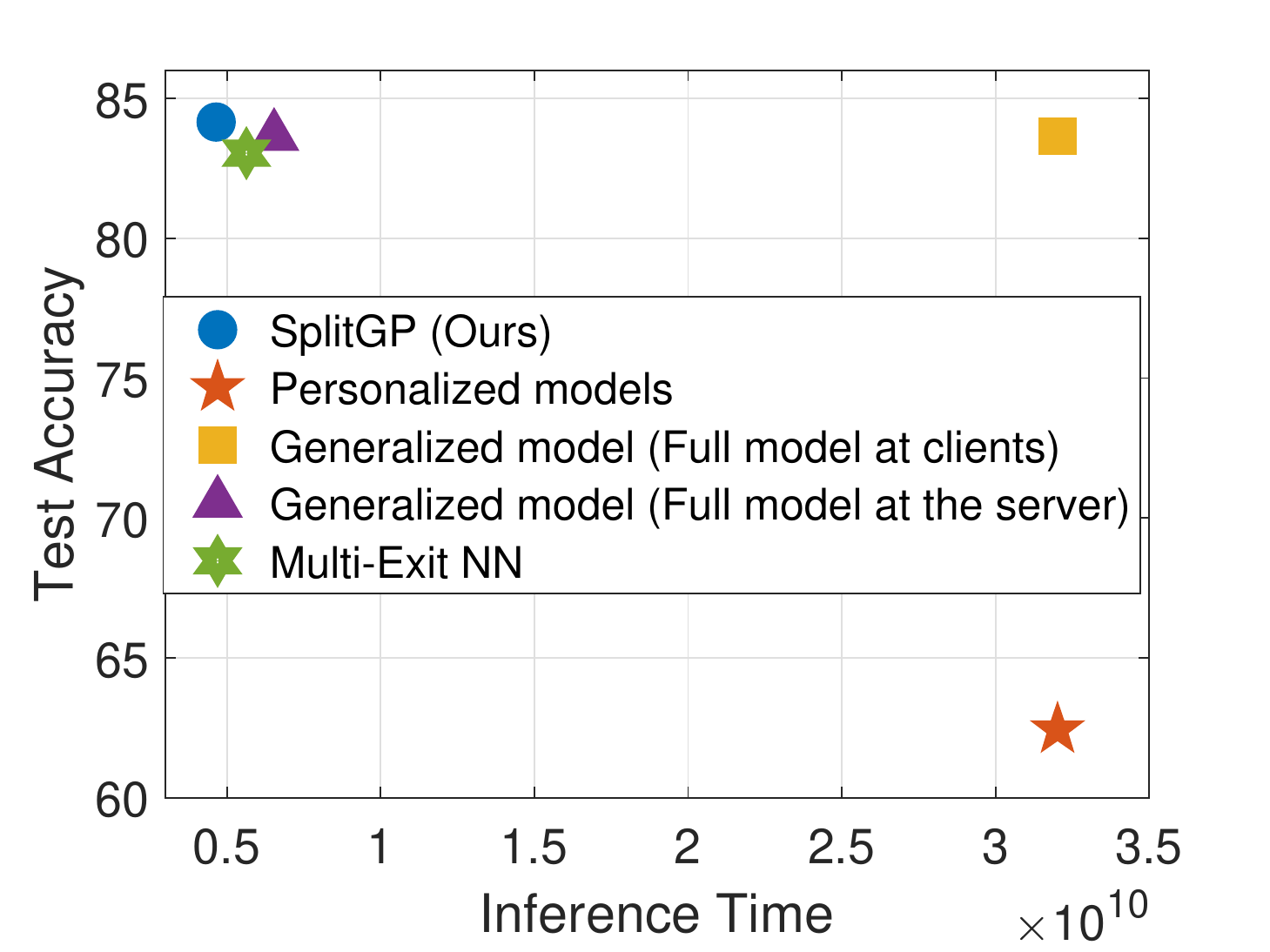}}
        \subfigure[CIFAR-10, $\rho=0.2$]{\includegraphics[width =0.24\textwidth]{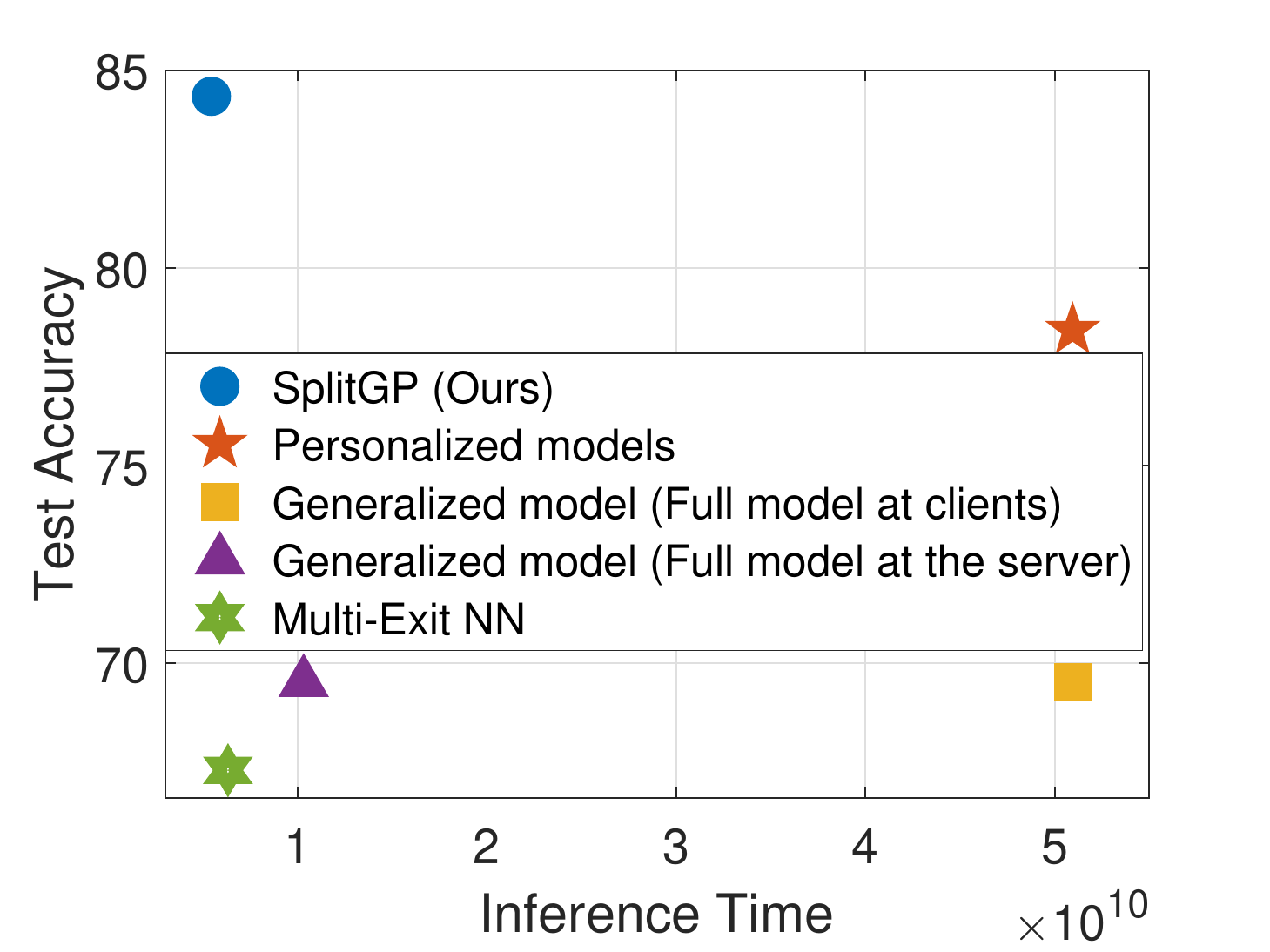}}
  \subfigure[CIFAR-10, $\rho=0.4$]{\includegraphics[width =0.24\textwidth]{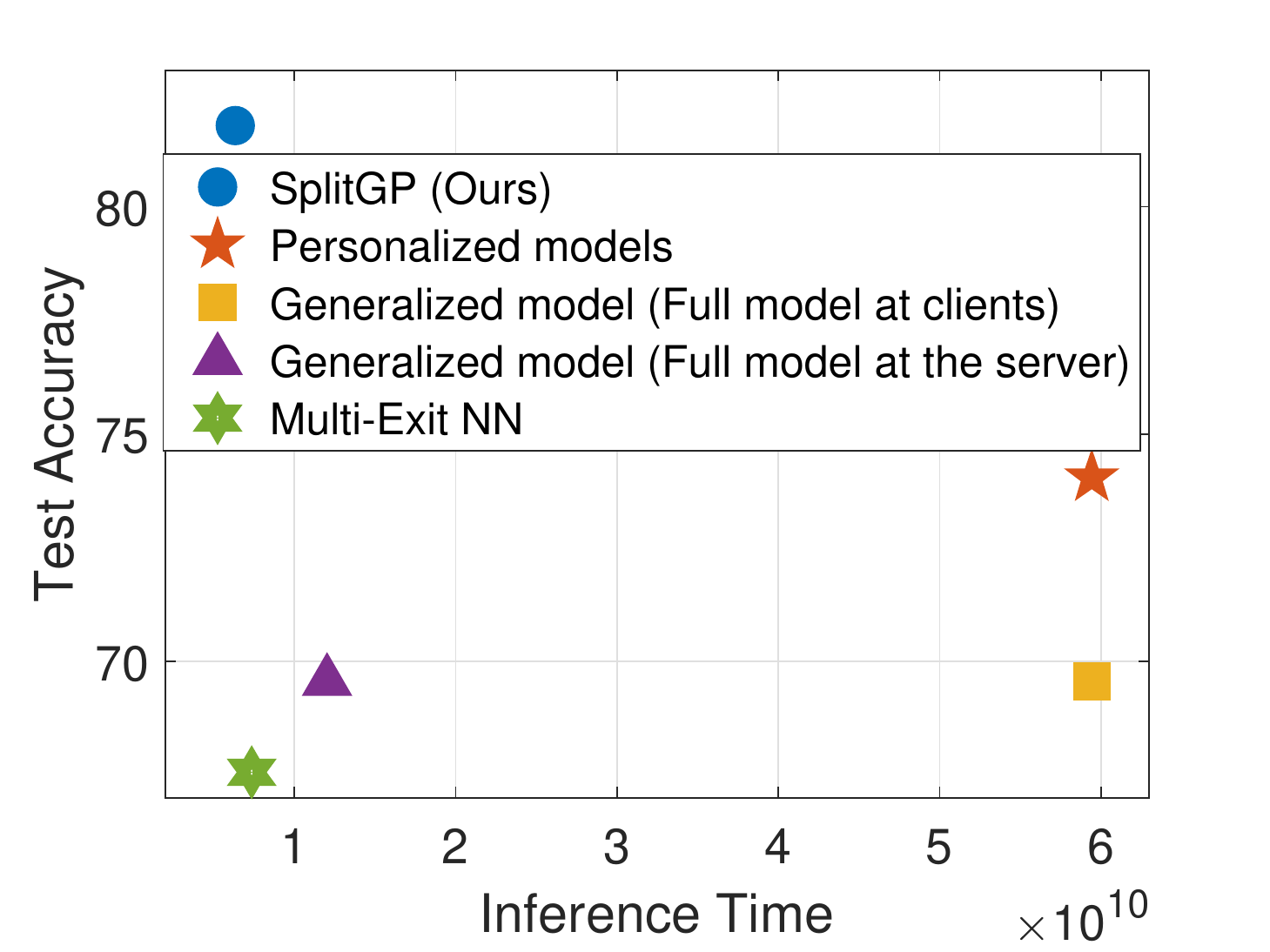}}
    \subfigure[CIFAR-10, $\rho=0.6$  ]{\includegraphics[width =0.24\textwidth]{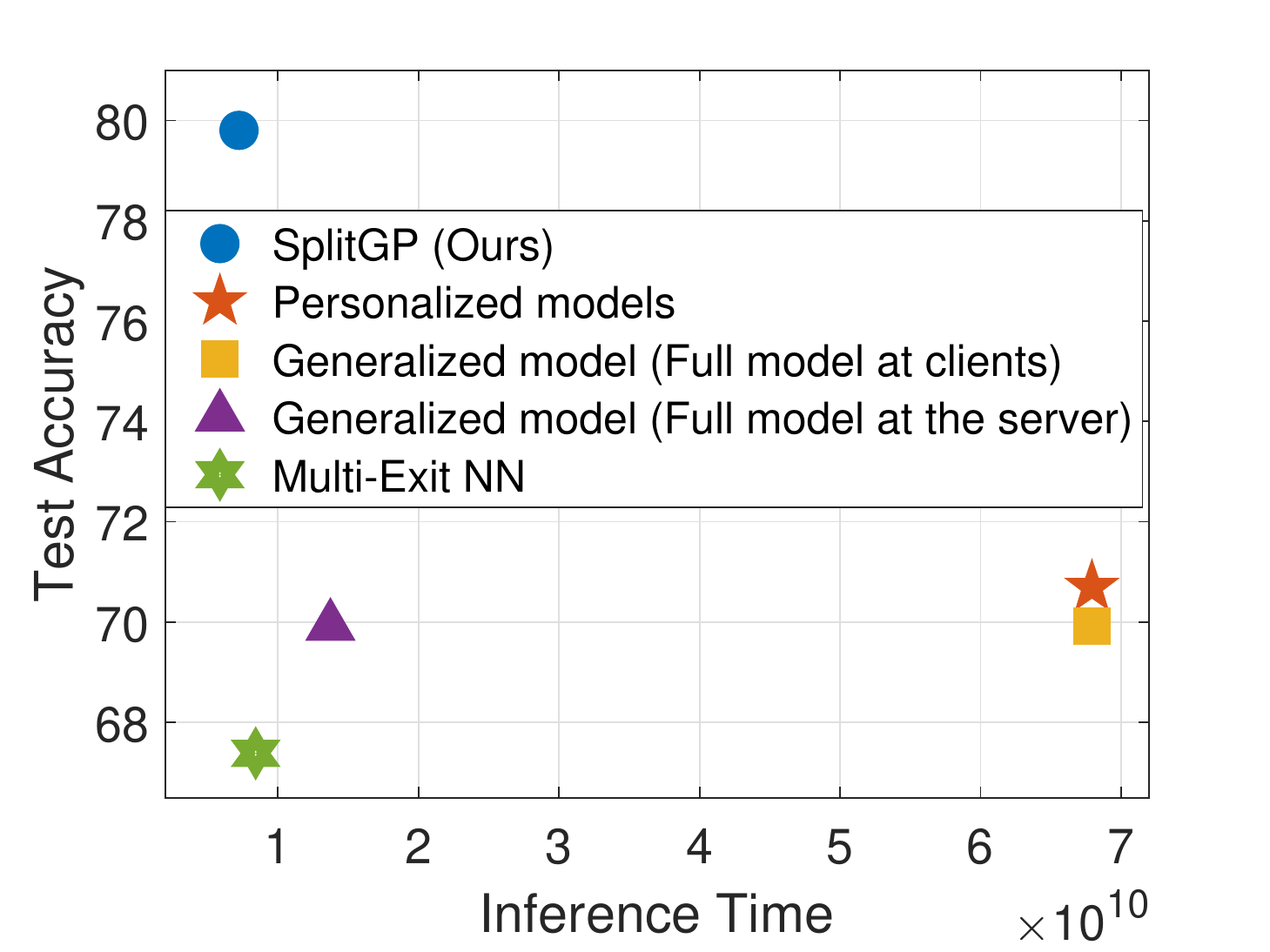}}
  \subfigure[CIFAR-10, $\rho=0.8$  ]{\includegraphics[width =0.24\textwidth]{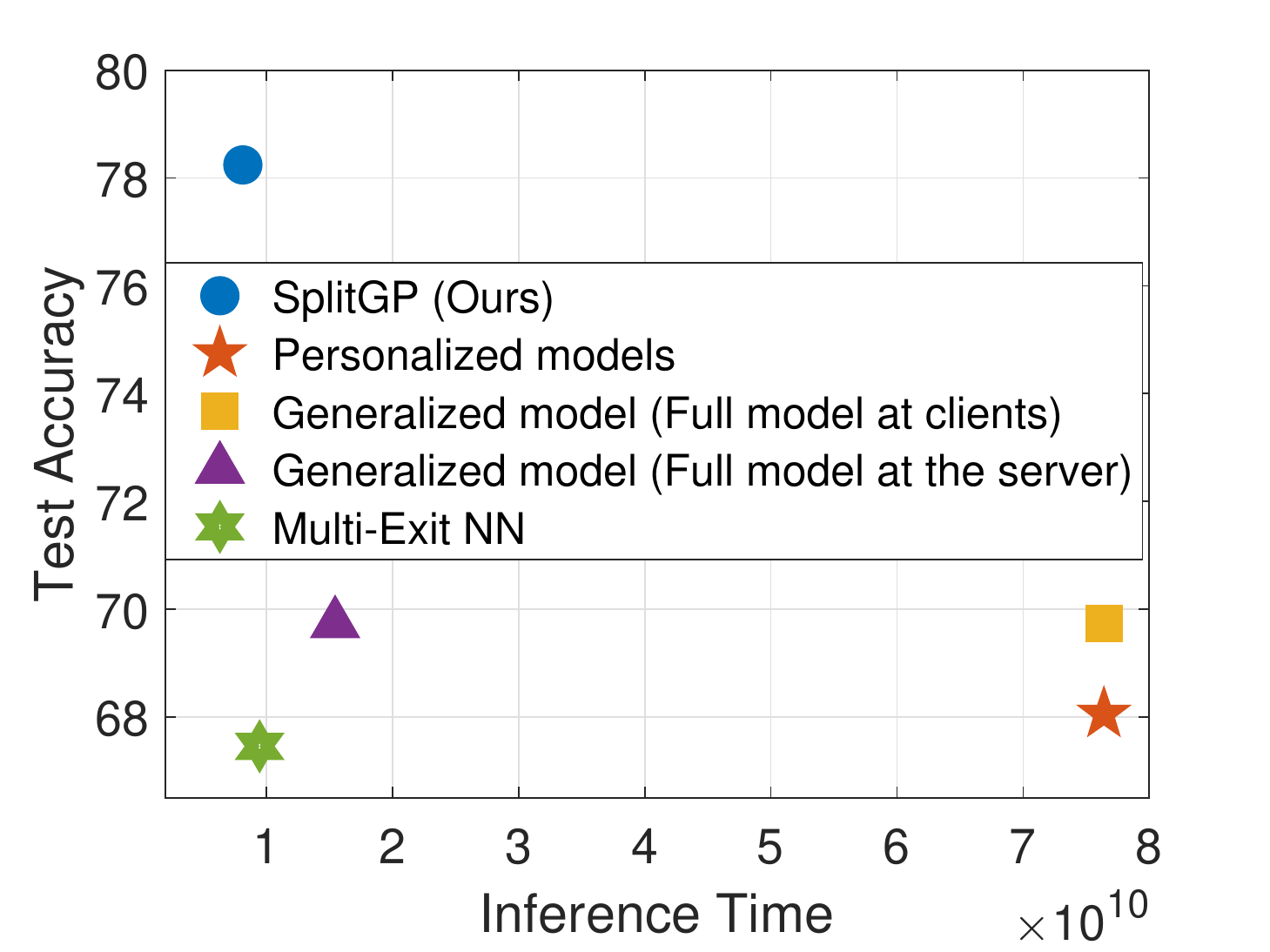}}
  \vspace{-3mm}
  \caption{ \small  Achievable accuracy-latency tradeoff. Our scheme achieves the best accuracy with smallest inference time for most settings of $\rho$ on both datasets, underscoring the ability of SplitGP to provide personalization and generalization while reducing inference resource requirements. } \label{fig:accuracy_vs_latency}
  \vspace{-3mm}
\end{figure*}




\textbf{Main result 1: Effect of out-of-distribution data.} We first observe Fig. \ref{fig:accuracy_vs_ratio}  and Table \ref{table:main_1}, which show  the performance of each scheme depending on the relative portion of out-of-distribution data $\rho$ during inference. 
We have the following key observations. First, the performance of the generalized global model and the  multi-exit neural network constructed via FL/SL do not dramatically change  with varying $\rho$. This implies that all classes pose a similar level of difficulty for classification, which is consistent with the class-balanced nature of    FMNIST and CIFAR-10.   It can be also seen that the performance of personalized FL is significantly degraded as $\rho$  grows, since   personalized models are designed to improve the   performance on the  main classes, not the out-of-distribution classes. Finally,  it is observed that   SplitGP   captures both personalization and generalization capabilities: due to the personalization capability,  our scheme achieves a strong performance when $\rho$ is small, and due  to the generalization capability, our scheme is more robust against    $\rho$   compared to   personalized FL.  



\textbf{Main result 2: Latency, accuracy, and resource improvements.} Fig. \ref{fig:accuracy_vs_latency}  shows the achievable accuracy-latency performance of the different schemes.  For   personalized FL,  the models are deployed at individual clients while the  generalized global model  can be deployed either at the client-side or at the server-side.  To evaluate the inference time, we compute the latency from  Table \ref{table:cost_compare} by setting $P_C=20$, $P_S=100$, $R=1$, as in Fig. 3.      It can be seen that SplitGP  achieves the best accuracy with smallest inference time for most values of $\rho$,   confirming the advantage of our solution.  
Note that this performance advantage   is achieved with considerable storage savings at the clients; compared to the case where the full model $w=[\phi, \theta]$ is deployed at each client, our scheme  only requires  $10.62\%$ and $10.64\%$ of the storage space for FMNIST and CIFAR-10, respectively, by saving only the client-side component $\phi$. The communication load is also significantly reduced compared to others; for example, when $\rho=0.8$ in FMNIST, our scheme achieves the best performance while inferring only $20.30\%$ of the test samples at the server. 



\textbf{Ablation 1: Effect of $\lambda$ and $E_{th}$.} 
In Fig. \ref{fig:lambda_effect} and Table \ref{table:lambda_effect}, we study the effect of $\lambda$ which controls the weights for personalization and generalization.      When $\lambda$ is relatively large, the weight for the personalized client-side model increases, which leads to    stronger personalization. However, the performance   degrades as $\rho$ increases, since the scheme with large $\lambda$ lacks generalization capability. 
In general, the best $\lambda$ depends on the $\rho$ value.  Without prior information, i.e., assuming $\rho$ is uniform in the range of $[0,1]$,   $\lambda=0.2$ gives the best expected accuracy.  On the other hand, if we have prior knowledge that $\rho$ is uniform in  $[0,0.2]$, $\lambda=0.3$ is a better option.  
\begin{figure}[t]
\vspace{-4mm}
\centering
  \subfigure[Effect of $\lambda$.  ]{\includegraphics[width=0.233\textwidth]{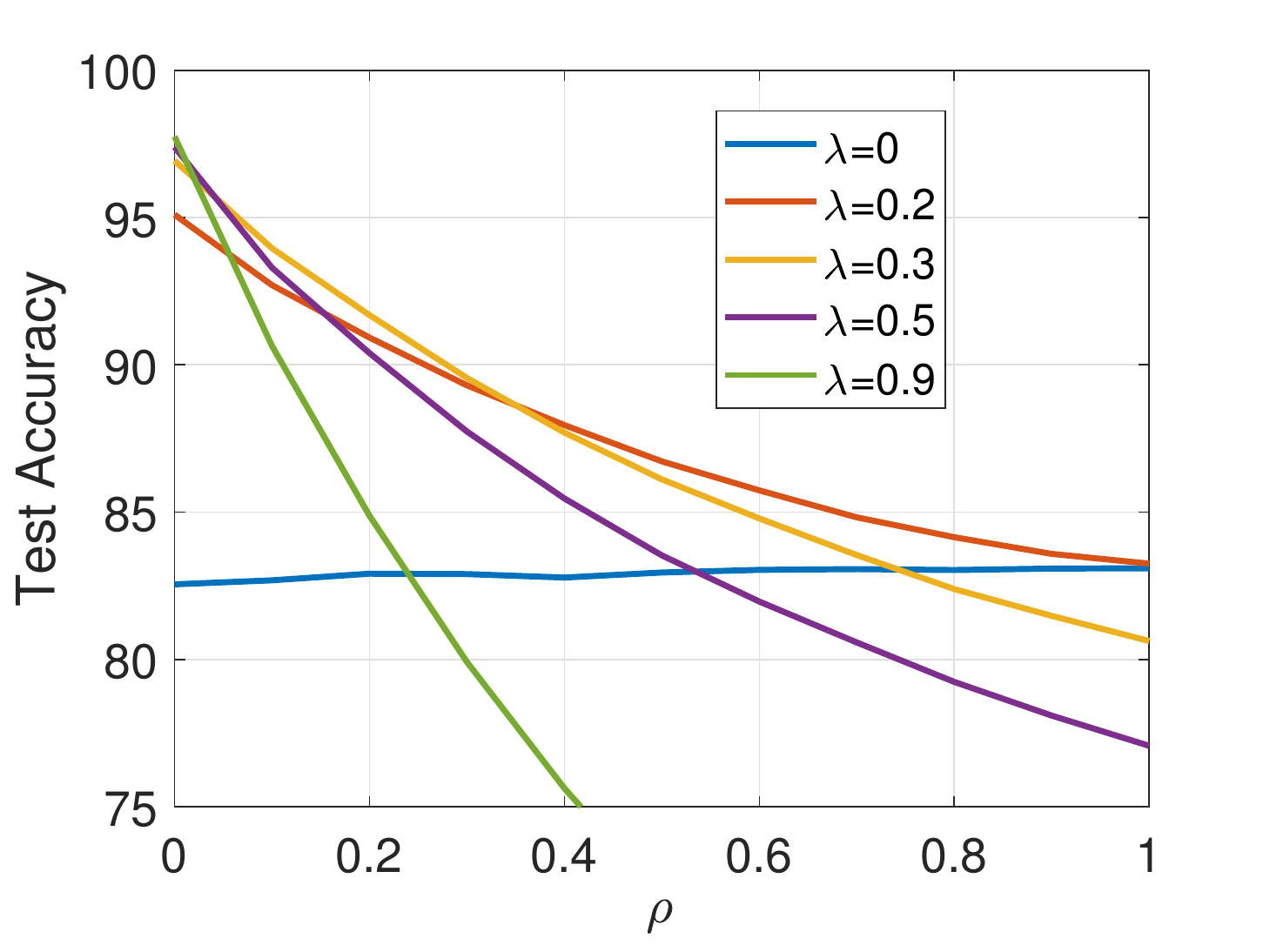}    \label{fig:lambda_effect}}
  \subfigure[Effect of  $E_{th}$. ]{\includegraphics[width =0.233\textwidth]{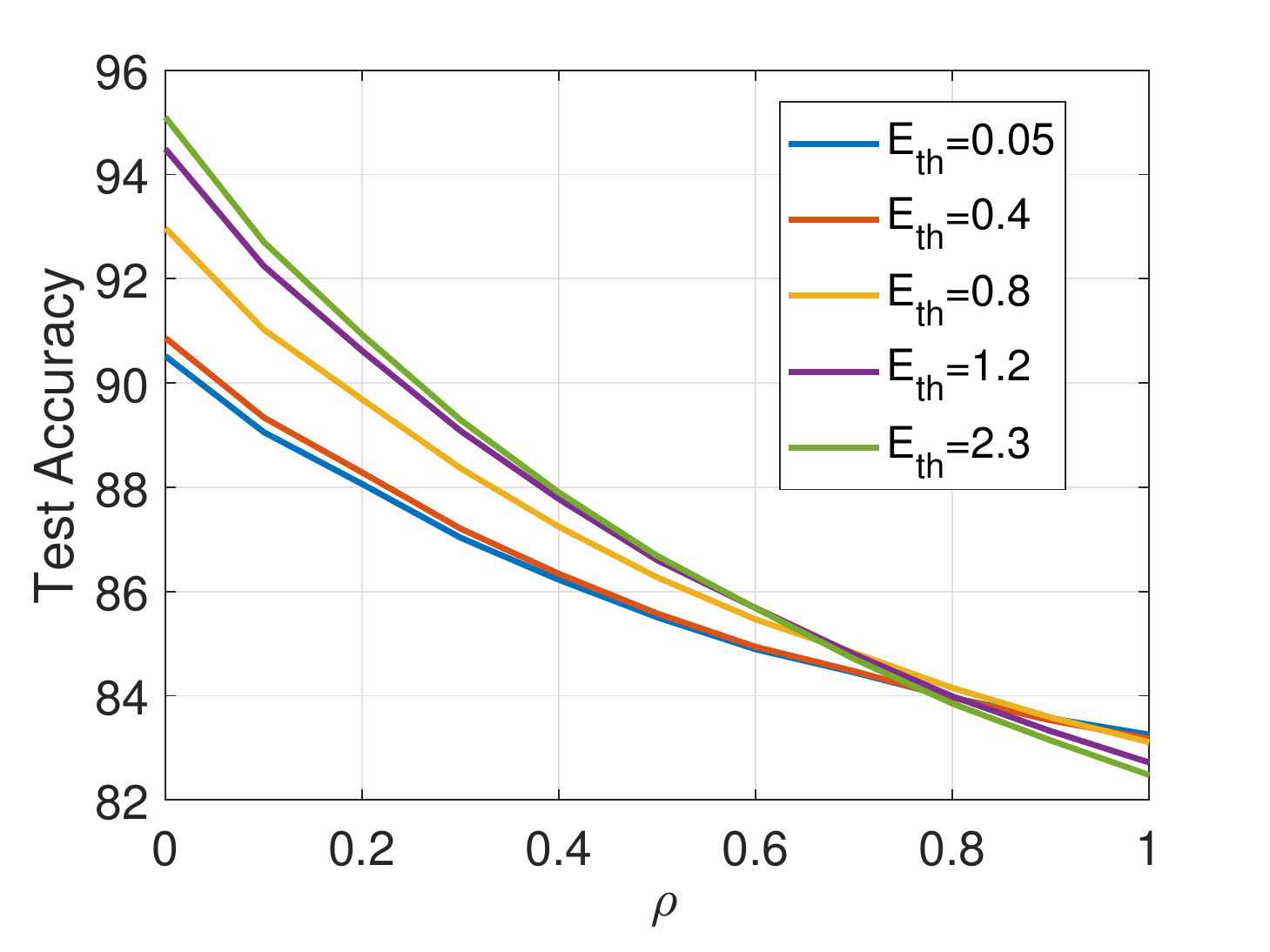}  \label{fig:accuracy_vs_Eth}}
  \vspace{-3mm} 
  \caption{\small Effects of $\lambda$ and $E_{th}$ in SplitGP for FMNIST. Larger $\lambda$ leads to   stronger personalization while smaller $\lambda$ leads to   stronger generalization.  A larger $E_{th}$ is a good option when $\rho$ is   small, while a smaller $E_{th}$ achieves a better performance when $\rho$ is large enough.}
  \vspace{-5mm}
\end{figure}

Now we observe the effect of $E_{th}$ in Fig. \ref{fig:accuracy_vs_Eth}. 
Similar to $\lambda$, one can   choose an appropriate $E_{th}$ given the  expected $\rho$ (or the range of $\rho$). 
When $\rho$ is small, a large $E_{th}$ performs well, which means that a relatively large number of samples should be predicted at the client-side to achieve the highest accuracy. On the other hand, when $\rho$ is large, smaller $E_{th}$ performs well which indicates that a large number of samples should be predicted at the server. These observations are consistent with our intuition that the main test samples should be predicted at the client-side (with strong personalization) while the out-of-distribution samples should be predicted at the server (with  strong generalization), to achieve the most robust performance.

\begin{table}
\vspace{-1mm}
\scriptsize
	\caption{\small  Effect of $\lambda$ on FMNIST. The value of  $\lambda$ should be chosen to achieve both generalization and personalization, depending on the expected  range of $\rho$.}
	\vspace{-1mm}
	\centering
	\label{table:lambda_effect}
	\begin{tabular}{l||ccccc}
		\toprule  
		\textbf{Methods} &  $\rho=0$ &$\rho=0.2$ &$\rho=0.4$ &$\rho=0.6$&$\rho=0.8$ \\ 
		\midrule
		$\lambda=0.2$&   $95.10\%$ & $90.93\%$ & $\mathbf{87.95}\%$ & $\mathbf{85.74}\%$ & $\mathbf{84.15}\%$\\
		$\lambda=0.3$&   $96.93\%$ & $\mathbf{91.69}\%$ & $87.70\%$ & $84.79\%$ & $82.39\%$\\
		$\lambda=0.5$&  $97.39\%$ & $90.40\%$ & $85.46\%$ & $81.96\%$ & $79.24\%$\\
		$\lambda=0.9$&   $\mathbf{97.75}\%$ & $ 84.87\%$ & $ 75.62\%$ & $68.79\%$ & $63.46\%$\\
		\bottomrule
	\end{tabular}
	  \vspace{-1mm}
\end{table}
%
%

\textbf{Ablation 2: Performance of each component.} 
Finally, we consider the performance of different components of our model.  Table \ref{table:aaaa} compares the   performance of the client-side model ($\phi_k$ combined with $h_k$) and the full model ($\phi_k$ combined with $\theta$) with the complete SplitGP on FMNIST.  Due to the personalization capability, it can be seen that SplitGP relies on the client model  when $\rho$ is small. As $\rho$ increases, SplitGP relies on both the client model and the server model to achieve generalization and personalization jointly. 

\begin{table}
\scriptsize
	\caption{\small  Performance of the client-side model and the full model  on FMNIST. Our scheme takes the benefits of both models.}   
	\vspace{-1mm}
	\centering
	\label{table:aaaa}
	\begin{tabular}{l||cccc}
		\toprule  
		\textbf{Methods} &  $\rho=0.2$ &$\rho=0.4$ &$\rho=0.6$ &$\rho=0.8$ \\ 
		\midrule
		Client  model (SplitGP)& $90.93\%$& 
  $87.90\%$ &  $85.68\%$ &$83.85\%$  \\ 
		Full model (SplitGP)& $88.06\%$ &$86.22\%$ &  $ 84.89\%$ &$83.96\%$ \\ 
		Overall performance (SplitGP)& $\mathbf{90.93}\%$  &   $\mathbf{87.95}\%$ &$\mathbf{85.74}\%$ & $\mathbf{84.15}\%$\\  
		\bottomrule
	\end{tabular}
	  \vspace{-4mm}
\end{table}




\vspace{-0.5mm}
\section{Conclusion}
\vspace{-0.5mm}
In this paper, we proposed a  hybrid federated and split learning methodology, termed SplitGP, which  captures  both personalization and generalization needs for reliable/efficient inference at resource-constrained clients. We analytically characterized     the convergence of our algorithm, and provided guidelines on model splitting based on  inference time analysis. Experimental results on real-world datasets confirmed the advantage of SplitGP in practical settings where each client needs  to make  predictions frequently for its main classes but also occasionally for its out-of-distribution classes.

\section*{Acknowledgement}
This work was supported by IITP funds from MSIT of Korea (No. 2020-0-00626, No. 2021-0-02201),  NRF (No. 2019R1I1A2A02061135, No. 2022R1A4A3033401), NSF CNS-2146171 and DARPA D22AP00168-00.  Minseok Choi is the corresponding author.

\bibliography{SL_INFOCOM_references}
\bibliographystyle{IEEEtran}

\end{document}